\documentclass{article} % For LaTeX2e

\usepackage[main, final]{neurips_2026}

\usepackage[utf8]{inputenc} % allow utf-8 input
\usepackage[T1]{fontenc}    % use 8-bit T1 fonts
\usepackage{hyperref}       % hyperlinks
\usepackage{url}            % simple URL typesetting
\usepackage{booktabs}       % professional-quality tables
\usepackage{amsfonts}       % blackboard math symbols
\usepackage{nicefrac}       % compact symbols for 1/2, etc.
\usepackage{microtype}      % microtypography
\usepackage{xcolor}         % colors
\usepackage{graphicx}
\usepackage{wrapfig}
\usepackage{array}
\usepackage{subcaption}
\usepackage{amsmath}
\usepackage{titletoc}
\usepackage{placeins}
\usepackage[normalem]{ulem}
\usepackage{makecell}
\title{Towards Financial World Modeling}%: \\ \vspace{2pt} \fontsize{12pt}{12pt}\selectfont The \texttt{Market-1T} Dataset and Training Effective Encoders for Financial Markets}

\author{
\hspace{-0.3cm}
\textbf{Humzah Merchant}$^{1}$,
\textbf{Alec Guthrie}$^{2}$,
\textbf{Simon Mahns}$^{3}$, 
\textbf{Randall Balestriero}$^{1,\dagger}$,
\textbf{Bradford Levy}$^{2,\dagger}$ \\
\hspace{-0.3cm}
$^{1}$Brown University \quad
$^{2}$University of Chicago \quad
$^{3}$Johns Hopkins University \quad
$^{\dagger}$Equal Advising \\
}

\begin{document}

\maketitle

\vspace{-1.5em}
\begin{abstract}
\vspace{-1em}
Building a world model requires a state representation useful for planning and decision-making---potentially over tasks unknown at training time. In the context of financial markets, planning and decision-making may require a model to reason about market-wide conditions, asset-specific expected returns, liquidity, volatility, and cross-asset relationships. Yet financial representation learning has largely been evaluated on individual predictive tasks, oftentimes on a single time period using comparatively narrow datasets. We address this through three primary contributions. First, we introduce \texttt{Market-1T}, a dataset containing nearly one trillion observations across U.S. equities from 2008 to 2025 at 1 Hz resolution. Second, we develop and implement a rigorous evaluation protocol. Third, we conduct a systematic large-scale study of financial representation learning, comparing 18 encoder-training strategies across nearly two decades of market regimes. We evaluate learned representations both by their predictive utility on common finance tasks and through probes of latent structure. We find that encoders with similar predictive performance can organize market state very differently. Collectively, we establish a foundation for training and evaluating financial market representations in support of world models such as DINO-WM, V-JEPA 2, and LeWM.\footnote{
Data, models, and code at
\href{https://huggingface.co/fin-ai-lab}{\textcolor{blue}{https://huggingface.co/fin-ai-lab}} and \href{https://github.com/fin-ai-lab/tfwm}{\textcolor{blue}{https://github.com/fin-ai-lab/tfwm}}. \texttt{bll@uchicago.edu}.
\vspace{-1em}
}
\end{abstract}

\begin{figure}[h]
    \centering
    \includegraphics[width=\linewidth]{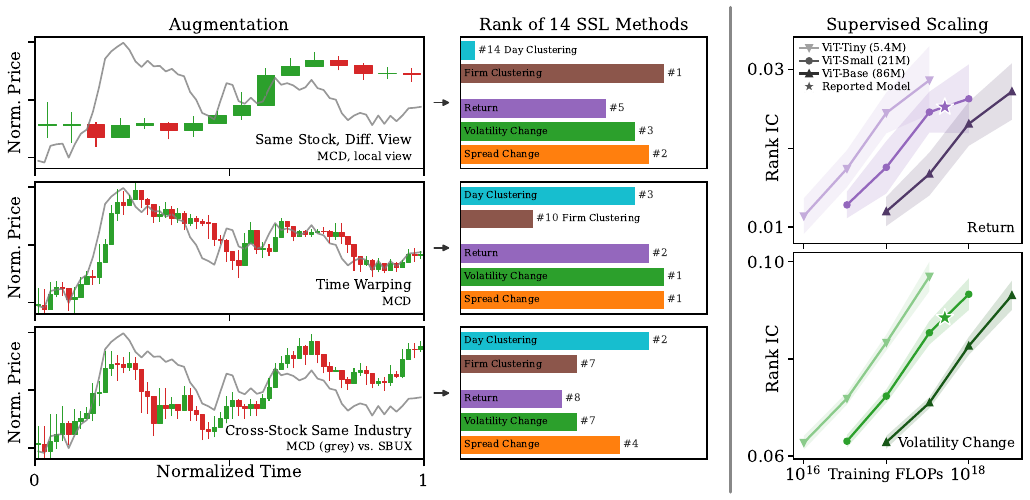}
    \caption{\textbf{Paper overview.} \textit{Left:} We systematically study how the pre-training objective and data augmentation shape the organization of the latent space and performance on downstream tasks such as return, volatility change, and spread change prediction. For example, matching views of the same stock at different times and resolutions yields latents that cluster tightly by firm but hardly at all by day, even though day-level structure, i.e., factor realizations, is arguably more important. \textit{Right:} On tasks such as return and volatility prediction, ViT-Tiny outperforms larger ViTs while fixing the number of training FLOPs. Further scaling results are presented in Figures~\ref{fig:scaling} and~\ref{fig:scaling_decade}.}
    \label{fig:hero}
\end{figure}

\newpage

\begin{figure}[h]
\vspace{-1em}
\centering
\includegraphics[width=\linewidth]{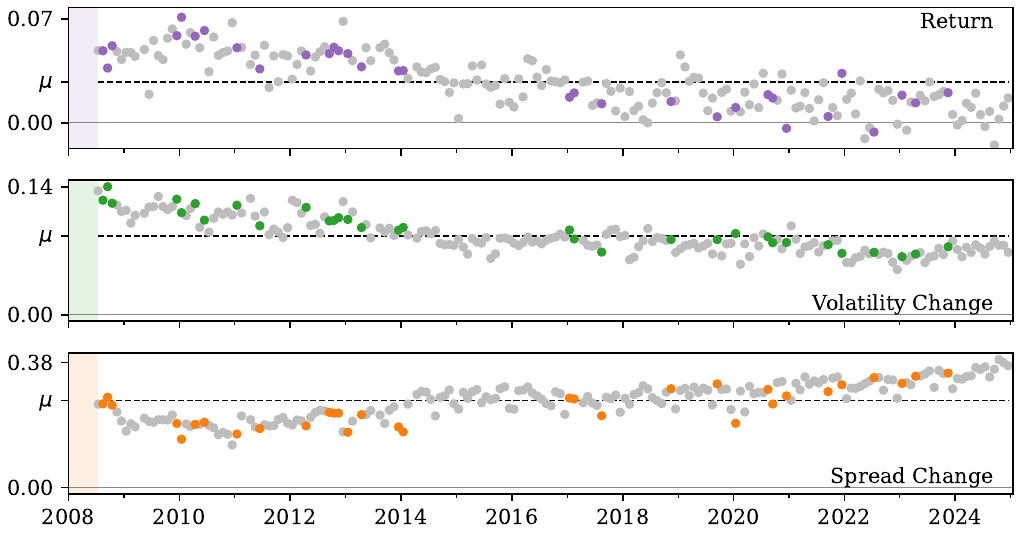}
\caption{Rank IC for the same encoder configuration and training setup across \texttt{Market-1T}'s 2008--2025 span. Performance varies substantially across market regimes, thus evaluation on narrow slices of time can lead to biased estimates of model performance with respect to it's true performance. We address this by using a randomly sampled set of 32 months.}
\label{fig:multihead_highlighted}
\vspace{-1em}
\end{figure}

\section{Introduction}

Financial markets play a key societal role in facilitating risk sharing \citep{arrow_1964}, providing liquidity \citep{ho_stoll_1981,glosten_milgrom_1985}, and aggregating information \citep{grossman_stiglitz_1980,kyle_1985, holmstrom_tirole_1993}. As such, the ability to monitor and understand markets then act accordingly is critical to maintaining a well-functioning economy and society \citep[see, e.g.,][]{clarida_etal_QJE_2000,orphanides,primiceri_QJE_2006,mavroeidis_AER_2010}. Key to such decision making is a model which can be used to explore counter-factual outcomes and optimize ex ante decisions, i.e., a ``world model.''

While financial markets are critical to modern economies, they are also extremely high-dimensional, noisy, geographically fragmented, and comprised of heterogeneous participants. For example, the state of a single asset over a day may involve millions of offers to buy or sell at particular prices and volumes across dozens of trading venues as well as trades by various participants---some human, some machine---with differing incentives. A world model for financial markets thus requires a state representation which enables planning and decision-making under a highly stochastic environment, varying market regimes, objectives, and tasks potentially unknown at training time.

While a financial world model likely needs to embed a rich state representation, representation learning for financial markets has historically focused on (i) comparatively narrow tasks known at training time and/or (ii) limited slices of markets---either temporally or asset-wise. We take the first step towards developing a world model for financial markets through three primary contributions:
\begin{enumerate}
    \item \textbf{Market-1T}: We introduce a publicly available and permissively licensed dataset of one-second quote and trade aggregates spanning all U.S.-traded equities from 2008 through 2025. With nearly one trillion observations over 18 years, \texttt{Market-1T} combines a longer historical span with substantially finer temporal resolution than is typical of public financial representation-learning benchmarks, making it possible to retrain and evaluate methods repeatedly across heterogeneous market regimes.
    \item \textbf{Rigorous Evaluation Protocol}: We closely examine the sensitivity of the evaluation protocol to probe sample size, market regime, model scale, representation age, and the portfolio-construction choices used to translate forecasts into trades, primarily explored with supervised baselines. We also compare these supervised baselines to common financial baselines, time series foundation models, and untrained encoders. Our final evaluation strategy samples 32 months throughout 2008--2025, with training restricted to preceding data and hyperparameters selected separately from the final evaluation periods.
    \item \textbf{Exploration of SSL in Financial Markets}: We compare 14 self-supervised encoder-training strategies, varying both data augmentation and pretraining objectives, under a common architecture and evaluation protocol. These encoders can form the foundation of LeWM \citep{leworldmodel2026}, DINO-WM \citep{zhou2025dinowm}, or V-JEPA-2 \citep{assran2025vjepa2selfsupervisedvideo} style world models. We ask both whether learning preserves information about future returns, volatility, and liquidity, and what economically meaningful structure the resulting latent spaces retain. 
\end{enumerate}

\textbf{The resulting picture is not captured by a single notion of representation quality.} Predictive performance varies substantially across market regimes, making long-horizon evaluation essential, while representations that perform well on forecasting need not be those that best preserve economically meaningful latent organization. These results provide both a benchmark for financial representation learning today and the state-representation layer for a broader program of financial world modeling: learning temporal and eventually action-conditioned dynamics on top of states that have been tested for what they preserve.

\begin{figure}[b]
    \centering
    \includegraphics[width=1\linewidth]{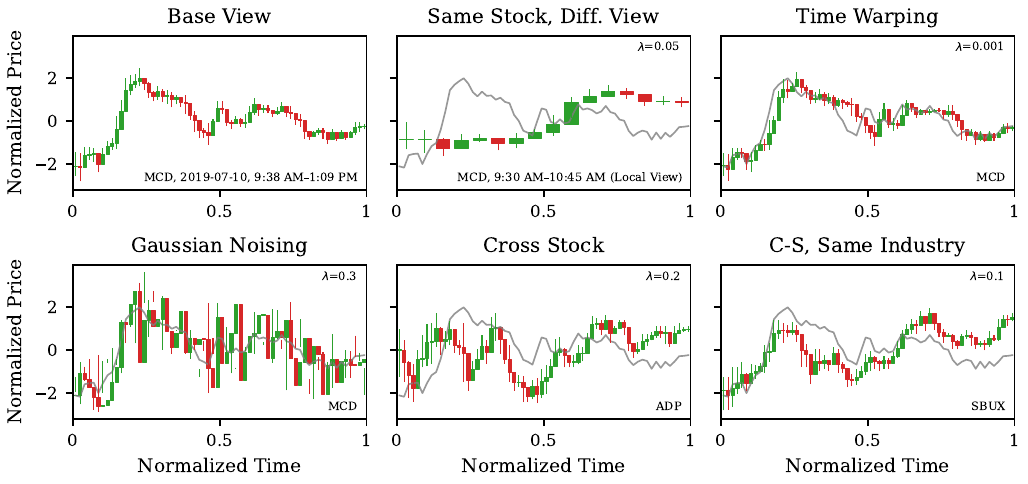}
    \caption{We evaluate five different augmentation strategies using LeJEPA. As shown in Tables \ref{tab:probe_fit} and \ref{tab:latent_results}, each augmentation learns different invariances about the market, e.g. same stock, diff. view primarily encourages memorization of firms through bid- and ask- size distributions, while cross-stock and cross-stock, same-industry emphasize more general market movements.}
    \label{fig:augmentations}
    \vspace{-1em}
\end{figure}

\section{Related Work}

\textbf{World modeling.} Modern world-modeling approaches differ in how the state representation is obtained: DINO-WM \citep{zhou2025dinowm} learns dynamics on top of a pretrained encoder, V-JEPA 2 \citep{assran2025vjepa2selfsupervisedvideo} adapts a pretrained predictive encoder to action-conditioned prediction, and LeWM \citep{leworldmodel2026} can learn or fine-tune the representation jointly with downstream dynamics. 
Our experiments focus on this representation layer: the financial encoders studied here can be inserted into these approaches as candidate state representations, either frozen or fine-tuned jointly with the dynamics model.

\textbf{Financial representation learning.} Financial representation-learning work has often been evaluated primarily through future prediction or through comparatively narrow downstream analyses. This includes general time-series forecasting models such as PatchTST \citep{nie2023timeseriesworth64} and Chronos \citep{ansari2024chronoslearninglanguagetime, ansari2025chronos2univariateuniversalforecasting}, high-frequency models such as DeepLOB \citep{Zhang_2019}, and finance-specific pretrained models such as DELPHYNE and Kronos \citep{ding2025delphynepretrainedmodelgeneral, shi2025kronosfoundationmodellanguage}. Existing self-supervised work has considered contrastive asset embeddings \citep{repec:arx:papers:2407.18645}, representations for stock similarity and portfolio construction \citep{repec:arx:papers:2407.13751}, features for portfolio diversification \citep{yang2023evaluation}, and reconstructive limit-order-book representations \citep{DBLP:journals/corr/abs-2406-19396}, and JEPA-based representations of latent market state \citep{mahns2026jointembedding}.
In contrast, we systematically compare a broad set of supervised, self-supervised, and pretrained encoders under a common multi-regime protocol, evaluating both information about future market evolution and economically meaningful structure in the learned latent space.

\textbf{Financial datasets and benchmarks.} Existing financial datasets vary substantially in scale, frequency, coverage, and availability. Public benchmarks include daily-frequency quantitative-equity datasets such as Qlib/Alpha158 \citep{yang2020qlibaiorientedquantitativeinvestment}, while high-frequency work often relies on specialized limit-order-book datasets and tasks, including those used by DeepLOB \citep{Zhang_2019} and SimLOB \citep{DBLP:journals/corr/abs-2406-19396}. Recent finance-specific pretrained models such as DELPHYNE and Kronos use substantially larger corpora \citep{ding2025delphynepretrainedmodelgeneral, shi2025kronosfoundationmodellanguage}, but these data are proprietary or only partially released. LOBSTER/TotalView is the closest in spirit to Market-1T, though it is organized at the event level and requires expensive licenses. Market-1T substantially expands the public setting along both the temporal and cross-sectional dimensions, containing nearly one trillion one-second aggregates across the changing universe of U.S.-traded equities from 2008--2025. Its combination of high temporal resolution, broad market coverage, and long historical span enables causal-in-time evaluation across many market regimes rather than evaluation on a fixed universe or short recent period. 

\section{Dataset}\label{sec:market1t}

\begin{table}[t]
\centering
\caption{Comparison of representative financial market datasets and corpora.}
\label{tab:dataset_comparison}
\resizebox{\linewidth}{!}{
\begin{tabular}{lllll}
\toprule
\textbf{Dataset} & \textbf{Span} & \textbf{Min. Resolution} & \textbf{Universe / Coverage} & \textbf{Access} \\
\midrule

\textbf{Market-1T}
& \textbf{2008--2025}
& \textbf{1 sec}
& \textbf{\begin{tabular}[c]{@{}l@{}}All U.S.-traded equities;\\consolidated SIP NBBO + trades\end{tabular}}
& \textbf{Public} \\ %$^{*}$

Kronos
& --
& 1 min
& \begin{tabular}[c]{@{}l@{}}45 exchanges; multiple\\asset classes\end{tabular}
& Partial \\

DELPHYNE
& --2019
& 5 min
& \begin{tabular}[c]{@{}l@{}}Global multi-asset; 15,817 securities\\in intraday-bar corpus\end{tabular}
& Proprietary \\ %$^{\ddagger}$

Qlib / \newline Alpha158
& 2008--2020
& Daily
& \begin{tabular}[c]{@{}l@{}}CSI300 Chinese A-shares\\(canonical benchmark)\end{tabular}
& Public \\

FI-2010
& 10 days (2010)
& Event-level
& \begin{tabular}[c]{@{}l@{}}5 Nasdaq Nordic stocks;\\10-level LOB\end{tabular}
& Public \\

LOBSTER / \newline TotalView
& 2007--present
& Event-level
& \begin{tabular}[c]{@{}l@{}}Nasdaq; reconstructed order book\\up to 200 levels\end{tabular}
& Licensed \\

\bottomrule
\end{tabular}
}
\vspace{-1em}

% {\footnotesize
% $^{*}$An 18-month sample is currently public; the full release is approved and in progress.
% \quad
% $^{\ddagger}$DELPHYNE also incorporates public LOTSA data; ``Proprietary'' refers to its finance-specific pretraining corpus.
% }
\end{table}

Building and evaluating financial world models requires data that capture both fine-grained market dynamics and the substantial distribution shifts that occur across market regimes. To support this, we partnered with \href{https://massive.com}{\texttt{Massive}} to construct \texttt{Market-1T}, a dataset spanning all equities traded on U.S. markets from 2008 through 2025. The universe includes both U.S. firms and non-U.S. firms whose equity is listed on U.S. exchanges. We begin in 2008, the first full year following the adoption of Regulation National Market System, which substantially changed U.S. trade-reporting requirements. Across the full sample, the dataset contains nearly one trillion one-second aggregates.

\textbf{Source and temporal resolution.}
\texttt{Market-1T} is constructed from consolidated U.S. equity quote and trade data distributed by \href{https://massive.com}{\texttt{Massive}}, including Securities Information Processor feeds from the Consolidated Tape Association (CTA) and Unlisted Trading Privileges (UTP) plans. The underlying data consist of National Best Bid and Offer (NBBO) updates and consolidated trade reports. We aggregate these asynchronous events into synchronous one-second observations.

We use 1 Hz as the base resolution because it retains fine-grained order-book and trading dynamics while permitting exact aggregation to coarser horizons. A model can therefore view the same underlying market history at resolutions ranging from seconds to minutes or longer without relying on interpolation. For each one-second interval, the order-book fields record the last valid quoted state available by the end of the interval, while trade fields summarize eligible transactions occurring during that interval.

Across resolutions, each observation contains the same 11 features:
\begin{itemize}
    \item \textbf{Order Book State:} best bid and ask prices and sizes
    ($P_{bid}, S_{bid}, P_{ask}, S_{ask}$).
    \item \textbf{Price Summary:} open, high, low, close, and volume-weighted average price (VWAP).
    \item \textbf{Activity:} traded volume ($V_{vol}$) and trade count ($N_{count}$).
\end{itemize}

\begin{figure}[t]
    \centering
    \includegraphics[width=1\linewidth]{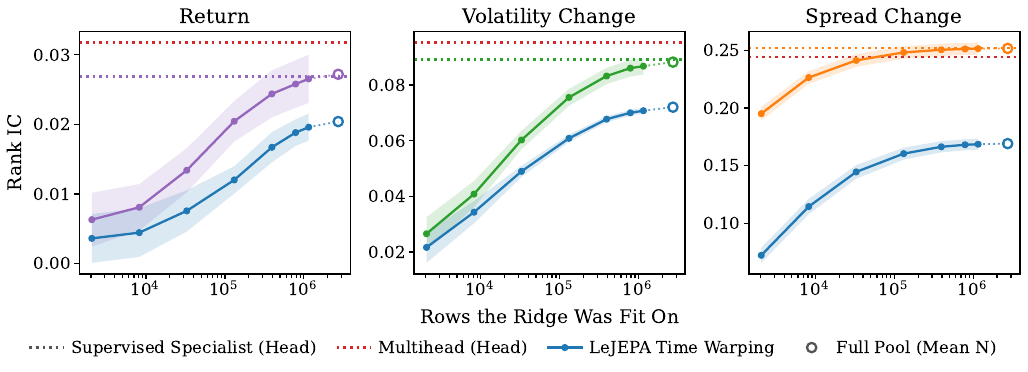}
    \caption{
    Predictive information retained by learned representations as a function of the number of samples used to fit the ridge probe. Probe performance improves substantially with sample size, indicating that large probe-training sets are required for reliable evaluation. LeJEPA \citep{balestriero2025lejepaprovablescalableselfsupervised} with time warping is among the strongest self-supervised methods. Full results are reported in Table~\ref{tab:probe_fit}.
    }
    \label{fig:main_prediction_results}
    \vspace{-1em}
\end{figure}

\textbf{Benchmark universe.}
Although \texttt{Market-1T} contains data for the full U.S.-traded equity universe, our benchmark experiments use a comparatively liquid common-stock universe. Crucially, the universe is recomputed separately for every month using only information available in the preceding month, avoiding survivorship and look-ahead bias. Securities must be common stocks, have prior-month average VWAP above \$5, average daily dollar volume of at least \$20M, and trading activity in more than 20\% of regular trading seconds. This produces a changing cross-section rather than a fixed list of firms that survived to the end of the sample. The released metadata support reproduction of this universe as well as construction of alternative universes.

\textbf{Aggregation for training.} The 1 Hz representation also allows us to construct views at different effective temporal scales using standard financial aggregation rules. For each output interval, we (i) take the last valid bid and ask state, (ii) compute the open, high, low, and close transaction prices, (iii) compute VWAP, and (iv) sum traded volume and number of trades. We use this operation in our random scaling and cropping augmentation, allowing different views of the same asset-day to span different amounts of clock time while preserving the semantics of market aggregates.

\textbf{Normalization during training.}
We normalize features according to their economic type. All price-related features are standardized jointly using a common mean and standard deviation, preserving their relative ordering within an observation (e.g., low, VWAP, high, bid, and ask). Bid and ask sizes are transformed using $\log(1+x)$ and then standardized jointly. Traded volume and trade count are separately transformed using $\log(1+x)$ and standardized. All normalization statistics are estimated from training data only.

Full details of data ingestion, event filtering, one-second aggregation, sparse storage, and universe construction are provided in Appendix~\ref{app_dataset_creation}.

\section{Evaluation}\label{sec:tasks}

\textbf{Protocol.} We evaluate every method on the same 32 months sampled across 2008–2025. For each evaluation month, models are trained using only the preceding six months (the Time Series Foundation Models are an exception to this rule.) We standardize architecture, compute, preprocessing, and probe fitting across methods wherever possible. Because probe quality is strongly sample-dependent (Figure \ref{fig:main_prediction_results}), all headline results use the full probe-training pool. We vary the pretraining objective and augmentations for SSL and use a standardized two stage procedure for hyper parameter optimization, performed on five separate tuning months. All plots present 2 standard error bars, with the SEs calculated adjusting for month-to-month variance (Figures \ref{fig:variance_decomp} and \ref{fig:variance_decomp_app}). Full training, sampling, augmentation, and hyperparameter-selection details are given in Appendix \ref{app:eval}.

\subsection{Predictive Information}

% requires \usepackage{xcolor}
\begin{table}[t]
\caption{Forecasting rank IC at the 15-minute horizon from a ridge probe, averaged over 32 eval months; a supervised arm's own head IC follows in parentheses, \textcolor{red!75!black}{red} cells fail to beat the untrained Random ViT floor, and \textcolor{green!45!black}{\textbf{green}} marks the best two \emph{non-supervised} arms per target. Finance baselines and TSFMs presented in Table \ref{tab:finance_baselines}.}
\label{tab:probe_fit}
\centering
\begin{tabular}{lccc}
\toprule
 & Return & Volatility & Spread \\
\midrule
\addlinespace \multicolumn{4}{l}{\emph{Supervised}} \\
Return & $0.0272$\,($0.0269$) & $0.0695$ & \textcolor{red!75!black}{$0.1305$} \\
Vol & $0.0197$ & $0.0883$\,($0.0891$) & \textcolor{red!75!black}{$0.1480$} \\
Spread & $0.0178$ & $0.0762$ & $0.2517$\,($0.2519$) \\
Multihead & $0.0311$\,($0.0317$) & $0.0941$\,($0.0952$) & $0.2455$\,($0.2441$) \\
\addlinespace \multicolumn{4}{l}{\emph{LeJEPA}~\citep{balestriero2025lejepaprovablescalableselfsupervised}} \\
\scriptsize{Same Stock, Diff. View} & $0.0182$ & $0.0715$ & \textcolor{red!75!black}{$0.1672$} \\
Time Warping & \textcolor{green!45!black}{$\mathbf{0.0204}$} & \textcolor{green!45!black}{$\mathbf{0.0721}$} & \textcolor{red!75!black}{$0.1691$} \\
Gaussian Noising & $0.0175$ & \textcolor{red!75!black}{$0.0688$} & \textcolor{red!75!black}{$0.1625$} \\
Cross Stock & \textcolor{red!75!black}{$0.0169$} & \textcolor{red!75!black}{$0.0693$} & \textcolor{red!75!black}{$0.1667$} \\
C-S, Same Industry & $0.0173$ & \textcolor{red!75!black}{$0.0692$} & \textcolor{red!75!black}{$0.1660$} \\
\addlinespace \multicolumn{4}{l}{\emph{SSL}} \\
DINO~\citep{caron_etal_dino} & \textcolor{red!75!black}{$0.0097$} & \textcolor{red!75!black}{$0.0310$} & \textcolor{red!75!black}{$0.0626$} \\
BYOL~\citep{grill2020bootstraplatentnewapproach} & \textcolor{red!75!black}{$0.0104$} & \textcolor{red!75!black}{$0.0314$} & \textcolor{red!75!black}{$0.0686$} \\
CPC~\citep{oord2019representationlearningcontrastivepredictive} & $0.0184$ & $0.0697$ & \textcolor{red!75!black}{$0.1519$} \\
I-JEPA~\citep{assran2023selfsupervisedlearningimagesjointembedding} & \textcolor{red!75!black}{$0.0103$} & \textcolor{red!75!black}{$0.0256$} & \textcolor{red!75!black}{$0.0646$} \\
MAE~\citep{he_etal_2022} & $0.0181$ & \textcolor{red!75!black}{$0.0684$} & \textcolor{red!75!black}{$0.1435$} \\
TS2Vec~\citep{yue2022ts2vec} & \textcolor{green!45!black}{$\mathbf{0.0205}$} & \textcolor{green!45!black}{$\mathbf{0.0719}$} & \textcolor{red!75!black}{$0.1500$} \\
CoST~\citep{woo2022costcontrastivelearningdisentangled} & \textcolor{red!75!black}{$0.0100$} & \textcolor{red!75!black}{$0.0339$} & \textcolor{red!75!black}{$0.0302$} \\
TF-C~\citep{zhang2022selfsupervisedcontrastivepretrainingtime} & \textcolor{red!75!black}{$0.0095$} & \textcolor{red!75!black}{$0.0570$} & \textcolor{red!75!black}{$0.1171$} \\
TimeMAE~\citep{Cheng_2026} & $0.0199$ & $0.0706$ & \textcolor{red!75!black}{$0.1408$} \\
\addlinespace \multicolumn{4}{l}{\emph{Untrained floor}} \\
Random ViT & $0.0170$ & $0.0694$ & $0.1692$ \\
\bottomrule
\end{tabular}
\vspace{-1em}
\end{table}

\begin{figure}
    \centering
    \includegraphics[width=1\linewidth]{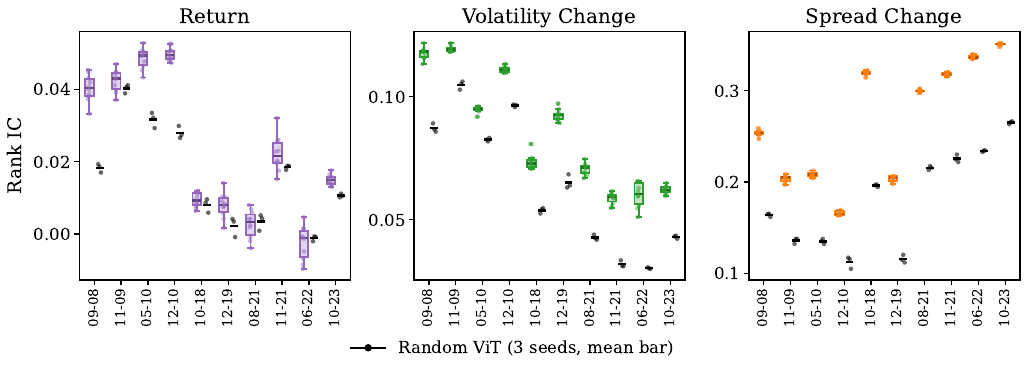}
    \caption{Decomposition of variance for our tasks between random seed and month ($10\times10$) along with the Random ViT baseline. We find that month contributes to 97.2\%, 98.9\%, and 99.9\% of variation for the three tasks, respectively. Therefore we subtract off the Random ViT mean when calculating standard errors, substantially reducing month-to-month variance (Figure \ref{fig:variance_decomp_app}).}
    \label{fig:variance_decomp}
    \vspace{-1em}
\end{figure}

Our first evaluation asks whether a learned representation preserves information about future market state. We probe representations for future return, volatility change \citep{andersen_etal_econometrica_2003}, and bid--ask spread change, the latter serving as a proxy for trading costs. We measure predictive information using rank information coefficient (IC), defining return as the change in volume-weighted average price (VWAP) from the minute after the view ends to the minute beginning 15 minutes after the view ends. Specifically:

\vspace{-1em}
\[
  \mathrm{IC} \;=\;\text{Spearman Correlation}\!\left(\hat{r}_{\cdot,t},\; \frac{\mathrm{VWAP}_{\cdot}(t{+}900,\,t{+}960)}{\mathrm{VWAP}_{\cdot}(t,\,t{+}60)} - 1\right)
\]
\vspace{-1em}

We average IC over all decision times in each evaluation month. We use rank IC as the primary predictive metric because it measures information in the forecast independently of the portfolio and execution choices used to monetize it. To study this, we evaluate 2,448 portfolio implementations varying universe selection, portfolio construction, risk modeling, rebalancing, and execution quality (Table \ref{tab:ic-sharpe-choices}). These choices move realized Sharpe substantially more than encoder choice, while IC remains strongly related to Sharpe after averaging over implementations (Figure \ref{fig:efq} and Table \ref{tab:ic-sharpe-harness}). \textbf{We further discuss the choice of IC vs Sharpe Ratio as a metric in Appendix~\ref{app:ic-vs-sharpe}.}

The core results are presented in Figure~\ref{fig:main_prediction_results} and Table~\ref{tab:probe_fit}. Supervised training preserves substantially more predictive information than the self-supervised objectives on several of the targets. Jointly supervised multihead training is particularly effective for return and volatility change, suggesting that these quantities share predictive structure that can be exploited by a common representation. Spread change behaves differently: the multihead model is slightly degraded from task spillover and all SSL methods perform worse than the untrained encoder, suggesting that information relevant to spread prediction is comparatively specialized.

The comparison to the untrained encoder is also important. A random ViT followed by a sufficiently well-fit probe is already a nontrivial baseline, and several self-supervised methods provide only modest improvements over it on the predictive tasks. Finally, predictive performance varies considerably more across evaluation periods than across random seeds. Figure~\ref{fig:variance_decomp} shows that month explains nearly all of the observed variation in evaluation performance. This makes evaluation across market regimes substantially more important than repeated evaluation within a single period and motivates the long-horizon protocol used throughout the paper.

\begin{figure}
    \centering
    \includegraphics[width=\linewidth]{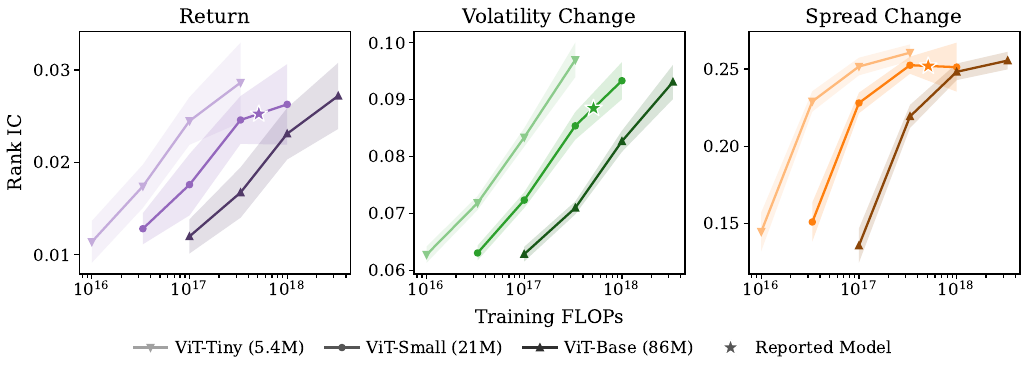}
    \caption{We find that for every task and every amount of FLOPs the smaller models appear a better fit. On all three tasks, the ViT-Tiny beats out a ViT-Base that has been trained to 10$\times$ the flops. We further expand the scaling in Figure \ref{fig:scaling_decade}.}
    \label{fig:scaling}
    \vspace{-1em}
\end{figure}

\subsubsection{Scaling and Decay}

We next study how predictive representations change with model scale and with the distance between the training and evaluation periods. At the scale considered here, these questions are more informative from a representation-learning perspective than from a practitioner perspective. The models are inexpensive to train relative to the scale at which a deployed forecasting system could be used, making frequent retraining and selection of an appropriately sized encoder operationally feasible. Thus, scaling and temporal decay are unlikely to be first-order deployment constraints in this setting. They are nevertheless useful for understanding the structure of the representation-learning problem.

At a fixed training-compute budget, the smaller encoders are consistently more effective across the predictive tasks. Increasing model size without a corresponding increase in useful statistical information therefore appears to reduce compute efficiency in this setting. For scaling we employ WSD \citep{hu2024minicpmunveilingpotentialsmall} and anneal all checkpoints. We present scaling for our default supervised configuration in Figure~\ref{fig:scaling} and a single run on the multihead, at up to 100$\times$ more FLOPs for the ViT-Tiny, in Figure \ref{fig:scaling_decade}.

As for temporal decay, because the intrinsic difficulty of the forecasting task changes across market regimes, directly evaluating performance as a function of model age would confound representation decay with changes in the predictability of the evaluation period. We therefore compare each model against an otherwise matched model trained closer to the evaluation date. Under this relative comparison, return representations appear comparatively persistent, while volatility and spread representations degrade as the training period moves further from evaluation.

\begin{figure}
    \centering
    \includegraphics[width=1\linewidth]{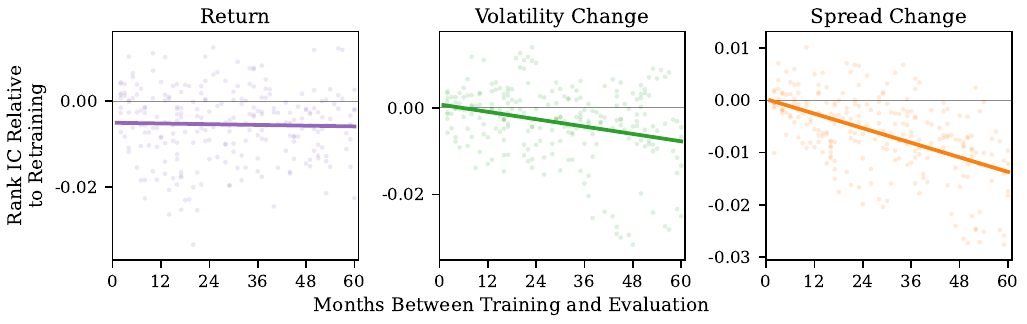}
    \caption{We evaluate decay by comparing models to the same model trained closer to the evaluation date to control for change in the intrinsic difficulty of the task over time. We see that it is small but present on volatility change and spread change, relative to the base IC ($\approx20\%$ of the base IC on volatility change and $\approx10\%$ on spread change, over five years.}
    \label{fig:forward_eval}
    \vspace{-1em}
\end{figure}

% requires \usepackage{xcolor}
\begin{table}[tp]
\centering
\includegraphics[width=1\linewidth]{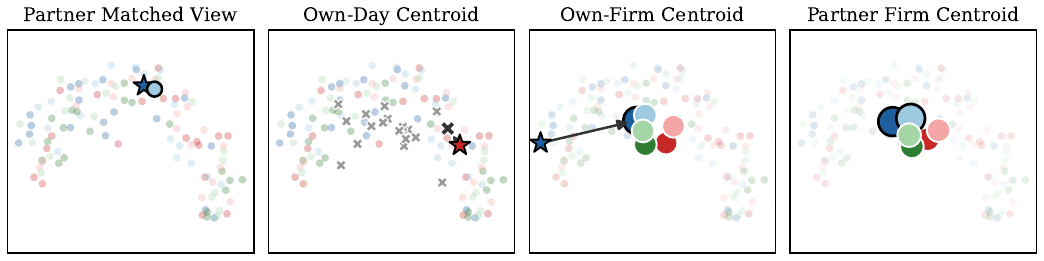}
\includegraphics[width=1\linewidth]{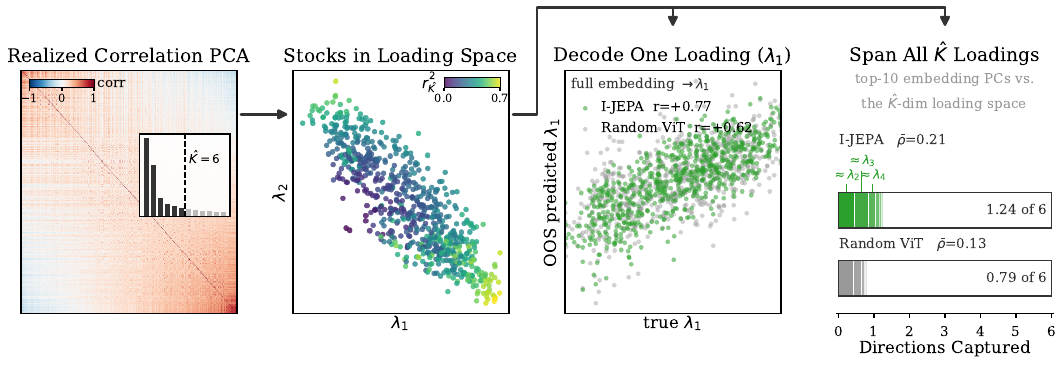}
\vspace{0.5em}
{\small
\setlength{\tabcolsep}{4pt}
\begin{tabular}{lcccccc}
\toprule
 & T1 & T2 & T3 & T4 & F1 & F2 \\
\midrule
\addlinespace \multicolumn{7}{l}{\emph{Supervised}} \\
Return & $2.26{\scriptstyle\times}$/$44.3{\scriptstyle\%}$ & $1.76{\scriptstyle\times}$/$42.8{\scriptstyle\%}$ & \textcolor{red!75!black}{$1.25{\scriptstyle\times}$}/\textcolor{red!75!black}{$45.6{\scriptstyle\%}$} & $1.17{\scriptstyle\times}$/\textcolor{red!75!black}{$47.3{\scriptstyle\%}$} & $0.633$ & $0.202$ \\
Vol & $2.13{\scriptstyle\times}$/$44.8{\scriptstyle\%}$ & $1.70{\scriptstyle\times}$/$43.2{\scriptstyle\%}$ & $1.91{\scriptstyle\times}$/$33.9{\scriptstyle\%}$ & $1.27{\scriptstyle\times}$/$45.6{\scriptstyle\%}$ & $0.581$ & $0.207$ \\
Spread & \textcolor{red!75!black}{$1.65{\scriptstyle\times}$}/\textcolor{red!75!black}{$46.3{\scriptstyle\%}$} & \textcolor{red!75!black}{$1.45{\scriptstyle\times}$}/\textcolor{red!75!black}{$45.5{\scriptstyle\%}$} & $2.41{\scriptstyle\times}$/$26.1{\scriptstyle\%}$ & \textcolor{red!75!black}{$1.15{\scriptstyle\times}$}/\textcolor{red!75!black}{$47.0{\scriptstyle\%}$} & \textcolor{red!75!black}{$0.459$} & $0.161$ \\
Multihead & $2.25{\scriptstyle\times}$/$44.4{\scriptstyle\%}$ & $1.70{\scriptstyle\times}$/$43.6{\scriptstyle\%}$ & $2.35{\scriptstyle\times}$/$28.6{\scriptstyle\%}$ & \textcolor{green!45!black}{$\mathbf{1.38{\scriptstyle\times}}$}/\textcolor{green!45!black}{$\mathbf{43.8{\scriptstyle\%}}$} & $0.582$ & \textcolor{green!45!black}{$\mathbf{0.243}$} \\
\addlinespace \multicolumn{7}{l}{\emph{LeJEPA}~\citep{balestriero2025lejepaprovablescalableselfsupervised}} \\
\scriptsize{Same Stock, Diff. View} & \textcolor{red!75!black}{$1.65{\scriptstyle\times}$}/$45.0{\scriptstyle\%}$ & \textcolor{red!75!black}{$1.31{\scriptstyle\times}$}/\textcolor{red!75!black}{$46.0{\scriptstyle\%}$} & \textcolor{green!45!black}{$\mathbf{5.36{\scriptstyle\times}}$}/\textcolor{green!45!black}{$\mathbf{2.5{\scriptstyle\%}}$} & $1.23{\scriptstyle\times}$/$45.5{\scriptstyle\%}$ & \textcolor{red!75!black}{$0.391$} & $0.199$ \\
Time Warping & $1.97{\scriptstyle\times}$/\textcolor{green!45!black}{$\mathbf{39.7{\scriptstyle\%}}$} & $2.01{\scriptstyle\times}$/\textcolor{green!45!black}{$\mathbf{37.7{\scriptstyle\%}}$} & \textcolor{red!75!black}{$1.54{\scriptstyle\times}$}/\textcolor{red!75!black}{$40.9{\scriptstyle\%}$} & \textcolor{red!75!black}{$1.07{\scriptstyle\times}$}/\textcolor{red!75!black}{$47.5{\scriptstyle\%}$} & $0.518$ & $0.218$ \\
Gaussian Noising & $2.33{\scriptstyle\times}$/$41.1{\scriptstyle\%}$ & $2.03{\scriptstyle\times}$/$39.2{\scriptstyle\%}$ & \textcolor{red!75!black}{$1.59{\scriptstyle\times}$}/\textcolor{red!75!black}{$38.6{\scriptstyle\%}$} & \textcolor{red!75!black}{$1.12{\scriptstyle\times}$}/\textcolor{red!75!black}{$47.3{\scriptstyle\%}$} & \textcolor{red!75!black}{$0.381$} & $0.162$ \\
Cross Stock & $2.46{\scriptstyle\times}$/$40.5{\scriptstyle\%}$ & $2.09{\scriptstyle\times}$/$38.4{\scriptstyle\%}$ & \textcolor{red!75!black}{$1.55{\scriptstyle\times}$}/\textcolor{red!75!black}{$39.3{\scriptstyle\%}$} & \textcolor{red!75!black}{$1.13{\scriptstyle\times}$}/\textcolor{red!75!black}{$47.6{\scriptstyle\%}$} & \textcolor{red!75!black}{$0.415$} & $0.167$ \\
C-S, Same Industry & \textcolor{green!45!black}{$\mathbf{2.58{\scriptstyle\times}}$}/$39.7{\scriptstyle\%}$ & \textcolor{green!45!black}{$\mathbf{2.22{\scriptstyle\times}}$}/\textcolor{green!45!black}{$\mathbf{37.6{\scriptstyle\%}}$} & \textcolor{red!75!black}{$1.52{\scriptstyle\times}$}/\textcolor{red!75!black}{$39.9{\scriptstyle\%}$} & \textcolor{red!75!black}{$1.12{\scriptstyle\times}$}/\textcolor{red!75!black}{$47.3{\scriptstyle\%}$} & \textcolor{red!75!black}{$0.433$} & $0.173$ \\
\addlinespace \multicolumn{7}{l}{\emph{SSL}} \\
DINO~\citep{caron_etal_dino} & $2.28{\scriptstyle\times}$/\textcolor{green!45!black}{$\mathbf{39.6{\scriptstyle\%}}$} & $2.04{\scriptstyle\times}$/$37.9{\scriptstyle\%}$ & \textcolor{red!75!black}{$1.49{\scriptstyle\times}$}/\textcolor{red!75!black}{$40.5{\scriptstyle\%}$} & $1.23{\scriptstyle\times}$/$46.1{\scriptstyle\%}$ & $0.544$ & $0.190$ \\
BYOL~\citep{grill2020bootstraplatentnewapproach} & $2.23{\scriptstyle\times}$/$41.0{\scriptstyle\%}$ & $1.94{\scriptstyle\times}$/$39.1{\scriptstyle\%}$ & \textcolor{red!75!black}{$1.72{\scriptstyle\times}$}/\textcolor{red!75!black}{$36.4{\scriptstyle\%}$} & $1.19{\scriptstyle\times}$/$46.4{\scriptstyle\%}$ & \textcolor{red!75!black}{$0.444$} & $0.199$ \\
CPC~\citep{oord2019representationlearningcontrastivepredictive} & $2.07{\scriptstyle\times}$/\textcolor{red!75!black}{$46.2{\scriptstyle\%}$} & $1.57{\scriptstyle\times}$/\textcolor{red!75!black}{$45.1{\scriptstyle\%}$} & \textcolor{red!75!black}{$1.55{\scriptstyle\times}$}/\textcolor{red!75!black}{$40.3{\scriptstyle\%}$} & \textcolor{red!75!black}{$1.15{\scriptstyle\times}$}/\textcolor{red!75!black}{$48.1{\scriptstyle\%}$} & \textcolor{red!75!black}{$0.475$} & $0.166$ \\
I-JEPA~\citep{assran2023selfsupervisedlearningimagesjointembedding} & $2.24{\scriptstyle\times}$/$40.5{\scriptstyle\%}$ & $1.89{\scriptstyle\times}$/$39.5{\scriptstyle\%}$ & \textcolor{red!75!black}{$1.06{\scriptstyle\times}$}/\textcolor{red!75!black}{$49.0{\scriptstyle\%}$} & $1.24{\scriptstyle\times}$/$45.1{\scriptstyle\%}$ & $0.608$ & $0.203$ \\
MAE~\citep{he_etal_2022} & $2.42{\scriptstyle\times}$/$40.2{\scriptstyle\%}$ & $2.10{\scriptstyle\times}$/$38.5{\scriptstyle\%}$ & \textcolor{red!75!black}{$1.23{\scriptstyle\times}$}/\textcolor{red!75!black}{$45.4{\scriptstyle\%}$} & $1.22{\scriptstyle\times}$/\textcolor{red!75!black}{$46.7{\scriptstyle\%}$} & $0.547$ & $0.178$ \\
TS2Vec~\citep{yue2022ts2vec} & $1.86{\scriptstyle\times}$/$43.5{\scriptstyle\%}$ & \textcolor{red!75!black}{$1.47{\scriptstyle\times}$}/$44.4{\scriptstyle\%}$ & \textcolor{green!45!black}{$\mathbf{3.99{\scriptstyle\times}}$}/\textcolor{green!45!black}{$\mathbf{11.0{\scriptstyle\%}}$} & $1.37{\scriptstyle\times}$/\textcolor{green!45!black}{$\mathbf{43.2{\scriptstyle\%}}$} & \textcolor{red!75!black}{$0.464$} & \textcolor{green!45!black}{$\mathbf{0.245}$} \\
CoST~\citep{woo2022costcontrastivelearningdisentangled} & $2.13{\scriptstyle\times}$/$40.6{\scriptstyle\%}$ & $2.00{\scriptstyle\times}$/$38.1{\scriptstyle\%}$ & \textcolor{red!75!black}{$1.03{\scriptstyle\times}$}/\textcolor{red!75!black}{$49.6{\scriptstyle\%}$} & $1.17{\scriptstyle\times}$/\textcolor{red!75!black}{$47.4{\scriptstyle\%}$} & \textcolor{green!45!black}{$\mathbf{0.711}$} & $0.181$ \\
TF-C~\citep{zhang2022selfsupervisedcontrastivepretrainingtime} & $2.15{\scriptstyle\times}$/\textcolor{red!75!black}{$45.6{\scriptstyle\%}$} & $1.55{\scriptstyle\times}$/\textcolor{red!75!black}{$45.0{\scriptstyle\%}$} & \textcolor{green!45!black}{$\mathbf{4.14{\scriptstyle\times}}$}/\textcolor{green!45!black}{$\mathbf{10.6{\scriptstyle\%}}$} & \textcolor{green!45!black}{$\mathbf{1.39{\scriptstyle\times}}$}/$44.7{\scriptstyle\%}$ & $0.534$ & $0.221$ \\
TimeMAE~\citep{Cheng_2026} & \textcolor{green!45!black}{$\mathbf{3.35{\scriptstyle\times}}$}/\textcolor{green!45!black}{$\mathbf{38.6{\scriptstyle\%}}$} & \textcolor{green!45!black}{$\mathbf{2.66{\scriptstyle\times}}$}/\textcolor{green!45!black}{$\mathbf{36.0{\scriptstyle\%}}$} & \textcolor{red!75!black}{$1.32{\scriptstyle\times}$}/\textcolor{red!75!black}{$43.9{\scriptstyle\%}$} & $1.33{\scriptstyle\times}$/$44.6{\scriptstyle\%}$ & $0.646$ & $0.207$ \\
\addlinespace \multicolumn{7}{l}{\emph{Frozen TSFMs} (last layer, channels averaged); see Figure~\ref{fig:tsfm_latent_sweep} for the layer-wise sweep.} \\
Chronos-2 & $2.24{\scriptstyle\times}$/$42.5{\scriptstyle\%}$ & $1.88{\scriptstyle\times}$/$39.9{\scriptstyle\%}$ & $2.56{\scriptstyle\times}$/$24.9{\scriptstyle\%}$ & \textcolor{red!75!black}{$1.07{\scriptstyle\times}$}/\textcolor{red!75!black}{$48.4{\scriptstyle\%}$} & \textcolor{green!45!black}{$\mathbf{0.674}$} & \textcolor{green!45!black}{$\mathbf{0.221}$} \\
Kronos & \textcolor{green!45!black}{$\mathbf{2.90{\scriptstyle\times}}$}/$41.1{\scriptstyle\%}$ & \textcolor{green!45!black}{$\mathbf{2.27{\scriptstyle\times}}$}/$38.6{\scriptstyle\%}$ & \textcolor{red!75!black}{$1.43{\scriptstyle\times}$}/\textcolor{red!75!black}{$42.2{\scriptstyle\%}$} & \textcolor{green!45!black}{$\mathbf{1.40{\scriptstyle\times}}$}/\textcolor{green!45!black}{$\mathbf{44.2{\scriptstyle\%}}$} & $0.620$ & $0.184$ \\
TimesFM 3.0 & $1.95{\scriptstyle\times}$/$44.2{\scriptstyle\%}$ & $1.55{\scriptstyle\times}$/$43.4{\scriptstyle\%}$ & $2.43{\scriptstyle\times}$/$28.2{\scriptstyle\%}$ & $1.22{\scriptstyle\times}$/$45.4{\scriptstyle\%}$ & \textcolor{green!45!black}{$\mathbf{0.716}$} & $0.194$ \\
\addlinespace \multicolumn{7}{l}{\emph{Untrained floor}} \\
Random ViT & $1.82{\scriptstyle\times}$/$45.6{\scriptstyle\%}$ & $1.48{\scriptstyle\times}$/$44.4{\scriptstyle\%}$ & $1.81{\scriptstyle\times}$/$36.2{\scriptstyle\%}$ & $1.16{\scriptstyle\times}$/$46.6{\scriptstyle\%}$ & $0.495$ & $0.151$ \\
\bottomrule
\end{tabular}
}

\vspace{-0.5em}

\caption{Evaluating the organization of learned market state. \textbf{Top:} We embed observations from six firms across three randomly selected industries within the same evaluation month and test whether proximity in representation space reflects shared day, firm, and industry structure, scored as T1--T4. \textbf{Bottom:} We estimate statistical factors from realized returns and measure both how well individual factor loadings can be decoded from the embeddings (F1) and how much of the factor subspace is captured by their principal directions (F2). \textbf{Table:} T1--T4 give the top-1 rate as a multiple of chance and the mean percentile rank of the target (lower is better), while F1 and F2 give a single score for which higher is better; \textcolor{green!45!black}{\textbf{green}} marks the top three trained methods in a column, and \textcolor{red!75!black}{red} values are no better than the untrained Random ViT.}
\label{tab:latent_results}
\end{table}

\subsection{Latent Structure}

Prediction alone does not determine whether a representation provides a useful state space for a world model. Spectral views of SSL interpret an objective as inducing relationships among observations and learning an embedding that is smooth over the resulting graph \citep{balestriero_spectral_graph_theory,balestriero2022contrastive}. In markets, such relationships arise from shared time periods, persistent firm characteristics, industry exposure, and common statistical factors, providing natural tests of whether a representation captures useful market structure beyond what is visible through next-return prediction alone.

\begin{figure}
    \centering
    \includegraphics[width=\linewidth]{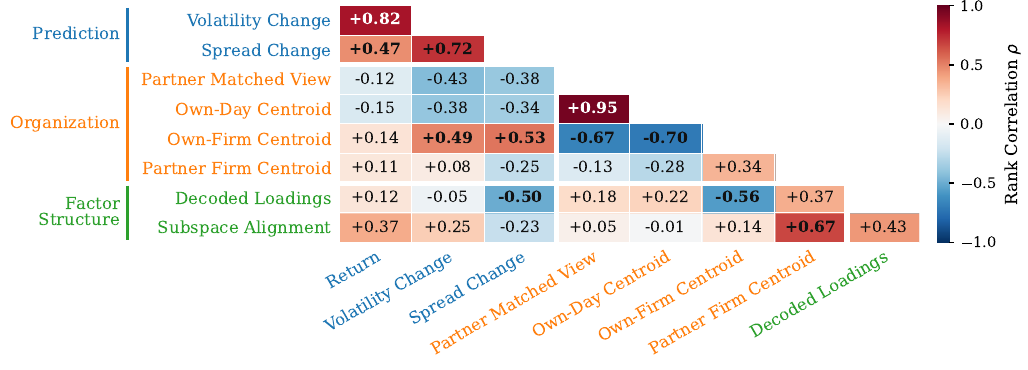}
    \includegraphics[width=\linewidth]{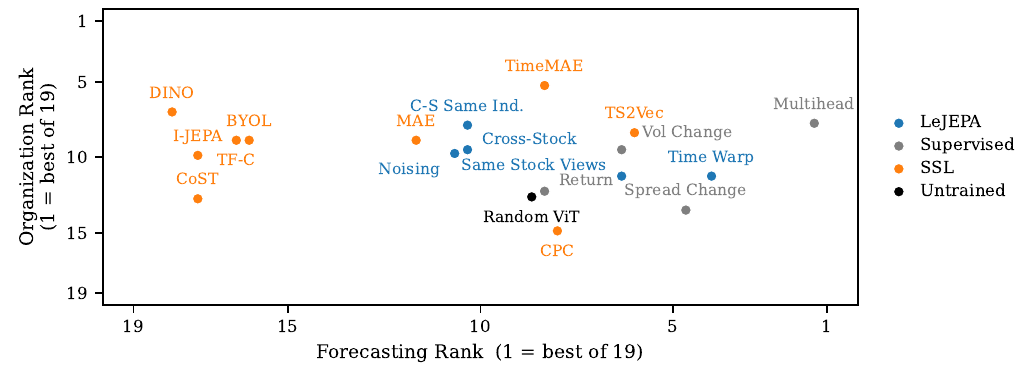}
    \caption{\textbf{Summary of Results:} We rank all 19 encoders (including the Random ViT) from 1 to 19 on each task and show the Spearman correlation between every pair of task rankings, with bold marking $p < 0.05$ (top). Averaging each encoder's ranks within the forecasting and organization tasks shows the two are unrelated ($\rho = -0.19$), with TimeMAE and Supervised Multihead drawing the efficient frontier.}
    \label{fig:summary}
\end{figure}

\textbf{Effective Rank.}
We first measure the effective rank of each representation, which is a useful characterization of how aggressively an objective compresses the observed market state, using RankMe \citep{garrido2023rankmeassessingdownstreamperformance}, with results shown in Table~\ref{tab:rankme_latent} and Figure~\ref{fig:rankme_sweep}. Supervised models tend to concentrate their variation into a comparatively small number of directions, while several self-supervised objectives produce substantially higher-dimensional embeddings.

\textbf{Organization.}
We next test whether distance in representation space reflects economically meaningful relationships between observations. A representation may organize observations according to the trading day, capturing market-wide conditions shared across firms, or according to firm identity and related cross-sectional characteristics, capturing persistent information about individual securities. We evaluate these alternatives through retrieval tasks involving matched views, day centroids, firm centroids, and economically related firms.

The resulting geometries differ substantially across objectives. Some methods focus on persistent firm identity while others focus on shared market state, as shown by the strong negative correlation between the `Own-Firm Centroid' and `Own-Day Centroid' in Figure \ref{fig:summary}. Some of the results are quite intuitive: for example, the model that is trained to see views of the same stock at different times learns to organize them using the bid and ask size information, whereas the models that take views of different stocks at the same time better learn to focus on general market movements while discarding information only useful for identifying a single stock. 

\textbf{Latent Statistical Factors.}
Our final latent-space evaluation asks whether the embeddings recover statistical factors that explain joint movements across stocks. Following \cite{pelger2018large}, we estimate factors from realized returns in the evaluation month. For each firm, we average its embeddings over the month and test two properties: (i) whether the firm’s loading on each return factor can be decoded from its embedding, and (ii) whether the leading principal components of the embeddings span the return-factor subspace. We cross-validate across groups of firms so that factor exposures are always evaluated on firms excluded from probe fitting. Unlike the preceding organization tasks, which define relatedness using firm, day, or industry labels, this evaluation defines relatedness directly from realized comovement in prices. This provides a complementary test of whether the representation preserves cross-asset dependence structure that could, for example, support covariance modeling.

\textbf{Summary.} Forecasting performance and latent-space organization capture largely distinct properties of a representation: across methods, their rankings are only weakly related ($\rho=-0.19$). Figure~\ref{fig:summary} summarizes these differences across the benchmark. The supervised multihead model is the only trained model that exceeds the Random ViT on every latent-space evaluation; among the pretrained models, TimesFM 3.0 does so as well (Table \ref{tab:latent_results}).

\section{Conclusion}

Monitoring and understanding financial markets, then acting to maximize expected outcomes is a central challenge in modern economies. Such decision making requires a world model of financial markets, however, building a world model begins with a representation-learning problem: before learning dynamics or planning actions, we need to know what information the state representation preserves and for which tasks that information is useful. In support of developing a financial world model, we introduce \texttt{Market-1T}, a nearly trillion-observation dataset spanning U.S.-traded equities from 2008--2025 at one-second resolution, develop a rigorous evaluation protocol, and use it to systematically evaluate 18 encoder-training strategies across multiple market regimes. We measure both predictive information about future market evolution and economically meaningful structure in the learned latent space.

Two conclusions emerge. First, predictive performance varies substantially across market regimes, making evaluation on a single recent period unreliable. Second, representation quality is multidimensional: methods that preserve predictive information need not preserve firm, temporal, cross-asset, or statistical-factor structure. This matters for world modeling because a model planning an action---for example, executing a trade---may need to reason jointly about expected returns, liquidity, volatility, market-wide conditions, and cross-asset relationships in order to anticipate both market evolution and the consequences of that action. A useful financial state representation must therefore preserve more than what is needed for next-return prediction alone.

At the time of writing, \href{https://massive.com}{\texttt{Massive}} is working on hosting an open and complete version of \texttt{Market-1T}. Code to assemble the full dataset manually at \href{https://github.com/fin-ai-lab/market-1t}{\textcolor{blue}{https://github.com/fin-ai-lab/market-1t}}.

\newpage
\section*{Funding Statement}

We acknowledge generous financial support from the Booth School of Business, the Center for Applied AI, and the Chookaszian Accounting Research Center. This research was supported in part by the Pythia computing cluster at The University of Chicago Booth School of Business which is funded by the Office of the Dean. The dataset released is in partnership with
\href{https://massive.com}{\texttt{Massive}}.
\par
\bibliographystyle{plain} % or plainnat, unsrt, abbrvnat, etc.
\bibliography{references} % assumes references.bib

\begin{thebibliography}{10}

\bibitem{andersen_etal_econometrica_2003}
Torben~G. Andersen, Tim Bollerslev, Francis~X. Diebold, and Paul Labys.
\newblock Modeling and forecasting realized volatility.
\newblock {\em Econometrica}, 71(2):579--625, 2003.

\bibitem{ansari2025chronos2univariateuniversalforecasting}
Abdul~Fatir Ansari, Oleksandr Shchur, Jaris Küken, Andreas Auer, Boran Han, Pedro Mercado, Syama~Sundar Rangapuram, Huibin Shen, Lorenzo Stella, Xiyuan Zhang, Mononito Goswami, Shubham Kapoor, Danielle~C. Maddix, Pablo Guerron, Tony Hu, Junming Yin, Nick Erickson, Prateek~Mutalik Desai, Hao Wang, Huzefa Rangwala, George Karypis, Yuyang Wang, and Michael Bohlke-Schneider.
\newblock Chronos-2: From univariate to universal forecasting, 2025.

\bibitem{ansari2024chronoslearninglanguagetime}
Abdul~Fatir Ansari, Lorenzo Stella, Caner Turkmen, Xiyuan Zhang, Pedro Mercado, Huibin Shen, Oleksandr Shchur, Syama~Sundar Rangapuram, Sebastian~Pineda Arango, Shubham Kapoor, Jasper Zschiegner, Danielle~C. Maddix, Hao Wang, Michael~W. Mahoney, Kari Torkkola, Andrew~Gordon Wilson, Michael Bohlke-Schneider, and Yuyang Wang.
\newblock Chronos: Learning the language of time series, 2024.

\bibitem{arrow_1964}
K.~J. Arrow.
\newblock The role of securities in the optimal allocation of risk-bearing.
\newblock {\em The Review of Economic Studies}, 31(2):91--96, 1964.

\bibitem{assran2023selfsupervisedlearningimagesjointembedding}
Mahmoud Assran, Quentin Duval, Ishan Misra, Piotr Bojanowski, Pascal Vincent, Michael Rabbat, Yann LeCun, and Nicolas Ballas.
\newblock Self-supervised learning from images with a joint-embedding predictive architecture, 2023.

\bibitem{assran2025vjepa2selfsupervisedvideo}
Mido Assran, Adrien Bardes, David Fan, Quentin Garrido, Russell Howes, Mojtaba, Komeili, Matthew Muckley, Ammar Rizvi, Claire Roberts, Koustuv Sinha, Artem Zholus, Sergio Arnaud, Abha Gejji, Ada Martin, Francois~Robert Hogan, Daniel Dugas, Piotr Bojanowski, Vasil Khalidov, Patrick Labatut, Francisco Massa, Marc Szafraniec, Kapil Krishnakumar, Yong Li, Xiaodong Ma, Sarath Chandar, Franziska Meier, Yann LeCun, Michael Rabbat, and Nicolas Ballas.
\newblock V-jepa 2: Self-supervised video models enable understanding, prediction and planning, 2025.

\bibitem{balestriero2022contrastive}
Randall Balestriero and Yann LeCun.
\newblock Contrastive and non-contrastive self-supervised learning recover global and local spectral embedding methods.
\newblock In Alice~H. Oh, Alekh Agarwal, Danielle Belgrave, and Kyunghyun Cho, editors, {\em Advances in Neural Information Processing Systems}, 2022.

\bibitem{balestriero2025lejepaprovablescalableselfsupervised}
Randall Balestriero and Yann LeCun.
\newblock Lejepa: Provable and scalable self-supervised learning without the heuristics, 2025.

\bibitem{balestriero_spectral_graph_theory}
Randall Balestriero and Yann LeCun.
\newblock Spectral graph theory: The mathematics of self-supervised learning [special issue on the mathematics of deep learning].
\newblock {\em IEEE Signal Processing Magazine}, 43(3):8--20, 2026.

\bibitem{caron_etal_dino}
Mathilde Caron, Hugo Touvron, Ishan Misra, Hervé Jegou, Julien Mairal, Piotr Bojanowski, and Armand Joulin.
\newblock Emerging properties in self-supervised vision transformers.
\newblock In {\em 2021 IEEE/CVF International Conference on Computer Vision (ICCV)}, pages 9630--9640, 2021.

\bibitem{Cheng_2026}
Mingyue Cheng, Xiaoyu Tao, Zhiding Liu, Qi~Liu, Hao Zhang, Rujiao Zhang, and Enhong Chen.
\newblock Timemae: Self-supervised representations of time series with decoupled masked autoencoders.
\newblock In {\em Proceedings of the Nineteenth ACM International Conference on Web Search and Data Mining}, page 498–508. ACM, February 2026.

\bibitem{clarida_etal_QJE_2000}
Richard Clarida, Jordi Galí, and Mark Gertler.
\newblock Monetary policy rules and macroeconomic stability: Evidence and some theory.
\newblock {\em The Quarterly Journal of Economics}, 115(1):147--180, 2000.

\bibitem{ding2025delphynepretrainedmodelgeneral}
Xueying Ding, Aakriti Mittal, and Achintya Gopal.
\newblock Delphyne: A pre-trained model for general and financial time series, 2025.

\bibitem{repec:arx:papers:2407.18645}
Rian Dolphin, Barry Smyth, and Ruihai Dong.
\newblock Contrastive learning of asset embeddings from financial time series.
\newblock Papers 2407.18645, arXiv.org, Jul 2024.

\bibitem{7104d0bd-69aa-330c-badb-94aa0c8f5bac}
Eugene~F. Fama.
\newblock Efficient capital markets: A review of theory and empirical work.
\newblock {\em The Journal of Finance}, 25(2):383--417, 1970.

\bibitem{FAMA1997153}
Eugene~F. Fama and Kenneth~R. French.
\newblock Industry costs of equity.
\newblock {\em Journal of Financial Economics}, 43(2):153--193, 1997.

\bibitem{garrido2023rankmeassessingdownstreamperformance}
Quentin Garrido, Randall Balestriero, Laurent Najman, and Yann Lecun.
\newblock Rankme: Assessing the downstream performance of pretrained self-supervised representations by their rank, 2023.

\bibitem{glosten_milgrom_1985}
Lawrence~R. Glosten and Paul~R. Milgrom.
\newblock Bid, ask and transaction prices in a specialist market with heterogeneously informed traders.
\newblock {\em Journal of Financial Economics}, 14(1):71--100, 1985.

\bibitem{grill2020bootstraplatentnewapproach}
Jean-Bastien Grill, Florian Strub, Florent Altché, Corentin Tallec, Pierre~H. Richemond, Elena Buchatskaya, Carl Doersch, Bernardo~Avila Pires, Zhaohan~Daniel Guo, Mohammad~Gheshlaghi Azar, Bilal Piot, Koray Kavukcuoglu, Rémi Munos, and Michal Valko.
\newblock Bootstrap your own latent: A new approach to self-supervised learning, 2020.

\bibitem{grossman_stiglitz_1980}
Sanford~J. Grossman and Joseph~E. Stiglitz.
\newblock On the impossibility of informationally efficient markets.
\newblock {\em The American Economic Review}, 70(3):393--408, 1980.

\bibitem{he_etal_2022}
Kaiming He, Xinlei Chen, Saining Xie, Yanghao Li, Piotr Dollár, and Ross Girshick.
\newblock Masked autoencoders are scalable vision learners.
\newblock In {\em 2022 IEEE/CVF Conference on Computer Vision and Pattern Recognition (CVPR)}, pages 15979--15988, 2022.

\bibitem{ho_stoll_1981}
Thomas Ho and Hans~R. Stoll.
\newblock Optimal dealer pricing under transactions and return uncertainty.
\newblock {\em Journal of Financial Economics}, 9(1):47--73, 1981.

\bibitem{holmstrom_tirole_1993}
Bengt Holmström and Jean Tirole.
\newblock Market liquidity and performance monitoring.
\newblock {\em Journal of Political Economy}, 101(4):678--709, 1993.

\bibitem{hu2024minicpmunveilingpotentialsmall}
Shengding Hu, Yuge Tu, Xu~Han, Chaoqun He, Ganqu Cui, Xiang Long, Zhi Zheng, Yewei Fang, Yuxiang Huang, Weilin Zhao, Xinrong Zhang, Zheng~Leng Thai, Kaihuo Zhang, Chongyi Wang, Yuan Yao, Chenyang Zhao, Jie Zhou, Jie Cai, Zhongwu Zhai, Ning Ding, Chao Jia, Guoyang Zeng, Dahai Li, Zhiyuan Liu, and Maosong Sun.
\newblock Minicpm: Unveiling the potential of small language models with scalable training strategies, 2024.

\bibitem{repec:arx:papers:2407.13751}
Yoontae Hwang, Stefan Zohren, and Yongjae Lee.
\newblock Temporal representation learning for stock similarities and its applications in investment management.
\newblock Papers 2407.13751, arXiv.org, Jul 2024.

\bibitem{kyle_1985}
Albert~S. Kyle.
\newblock Continuous auctions and insider trading.
\newblock {\em Econometrica}, 53(6):1315--1335, 1985.

\bibitem{LEDOIT2003603}
Olivier Ledoit and Michael Wolf.
\newblock Improved estimation of the covariance matrix of stock returns with an application to portfolio selection.
\newblock {\em Journal of Empirical Finance}, 10(5):603--621, 2003.

\bibitem{levy2022price}
Bradford~(Lynch) Levy.
\newblock Price improvement and payment for order flow: Evidence from a randomized controlled trial.
\newblock June 2022.
\newblock Jacobs Levy Equity Management Center for Quantitative Financial Research Paper.

\bibitem{DBLP:journals/corr/abs-2406-19396}
Yuanzhe Li, Yue Wu, and Peng Yang.
\newblock Simlob: Learning representations of limited order book for financial market simulation.
\newblock {\em CoRR}, abs/2406.19396, 2024.

\bibitem{leworldmodel2026}
Lucas Maes, Quentin Le~Lidec, Damien Scieur, Yann LeCun, and Randall Balestriero.
\newblock Leworldmodel: Stable end-to-end joint-embedding predictive architecture from pixels.
\newblock 2026.

\bibitem{mahns2026jointembedding}
Simon Mahns, Randall Balestriero, and Mahmoud Assran.
\newblock Joint-embedding predictive learning of latent market states in u.s. equities.
\newblock In {\em Forty-third International Conference on Machine Learning}, 2026.

\bibitem{maillard2010properties}
S{\'e}bastien Maillard, Thierry Roncalli, and J{\'e}r{\^o}me Teiletche.
\newblock On the properties of equally-weighted risk contributions portfolios.
\newblock {\em The Journal of Portfolio Management}, 36(4):60--70, 2010.

\bibitem{e5a1bb8f-41b7-35c6-95cd-8b366d3e99bc}
Harry Markowitz.
\newblock Portfolio selection.
\newblock {\em The Journal of Finance}, 7(1):77--91, 1952.

\bibitem{mavroeidis_AER_2010}
Sophocles Mavroeidis.
\newblock Monetary policy rules and macroeconomic stability: Some new evidence.
\newblock {\em American Economic Review}, 100(1):491–503, March 2010.

\bibitem{nie2023timeseriesworth64}
Yuqi Nie, Nam~H. Nguyen, Phanwadee Sinthong, and Jayant Kalagnanam.
\newblock A time series is worth 64 words: Long-term forecasting with transformers, 2023.

\bibitem{orphanides}
Athanasios Orphanides.
\newblock Monetary policy rules based on real-time data.
\newblock {\em American Economic Review}, 91(4):964–985, September 2001.

\bibitem{pelger2018large}
Markus Pelger.
\newblock Large-dimensional factor modeling based on high-frequency observations.
\newblock May 2018.
\newblock Available at SSRN.

\bibitem{primiceri_QJE_2006}
Giorgio~E. Primiceri.
\newblock Why inflation rose and fell: Policy-makers' beliefs and u. s. postwar stabilization policy*.
\newblock {\em The Quarterly Journal of Economics}, 121(3):867--901, 08 2006.

\bibitem{shi2025kronosfoundationmodellanguage}
Yu~Shi, Zongliang Fu, Shuo Chen, Bohan Zhao, Wei Xu, Changshui Zhang, and Jian Li.
\newblock Kronos: A foundation model for the language of financial markets, 2025.

\bibitem{oord2019representationlearningcontrastivepredictive}
Aaron van~den Oord, Yazhe Li, and Oriol Vinyals.
\newblock Representation learning with contrastive predictive coding, 2019.

\bibitem{woo2022costcontrastivelearningdisentangled}
Gerald Woo, Chenghao Liu, Doyen Sahoo, Akshat Kumar, and Steven Hoi.
\newblock Cost: Contrastive learning of disentangled seasonal-trend representations for time series forecasting, 2022.

\bibitem{yang2020qlibaiorientedquantitativeinvestment}
Xiao Yang, Weiqing Liu, Dong Zhou, Jiang Bian, and Tie-Yan Liu.
\newblock Qlib: An ai-oriented quantitative investment platform, 2020.

\bibitem{yang2023evaluation}
Yongxin Yang and Timothy~M. Hospedales.
\newblock An evaluation of self-supervised learning for portfolio diversification.
\newblock In {\em International Conference on Artificial Neural Networks}, 2023.
\newblock Available at SSRN: \url{https://ssrn.com/abstract=4187326}.

\bibitem{yue2022ts2vec}
Zhihan Yue, Yujing Wang, Juanyong Duan, Tianmeng Yang, Congrui Huang, Yunhai Tong, and Bixiong Xu.
\newblock Ts2vec: Towards universal representation of time series.
\newblock {\em Proceedings of the AAAI Conference on Artificial Intelligence}, 36(8):8980--8987, 2022.

\bibitem{zhang2022selfsupervisedcontrastivepretrainingtime}
Xiang Zhang, Ziyuan Zhao, Theodoros Tsiligkaridis, and Marinka Zitnik.
\newblock Self-supervised contrastive pre-training for time series via time-frequency consistency, 2022.

\bibitem{Zhang_2019}
Zihao Zhang, Stefan Zohren, and Stephen Roberts.
\newblock Deeplob: Deep convolutional neural networks for limit order books.
\newblock {\em IEEE Transactions on Signal Processing}, 67(11):3001–3012, 2019.

\bibitem{zhou2025dinowm}
Gaoyue Zhou, Hengkai Pan, Yann LeCun, and Lerrel Pinto.
\newblock {DINO}-{WM}: World models on pre-trained visual features enable zero-shot planning.
\newblock In {\em Forty-second International Conference on Machine Learning}, 2025.

\end{thebibliography}

\newpage
\appendix

\section*{Appendix Overview}
The appendix provides additional methodological details, supplementary analyses,
and discussion supporting the main paper. Appendix A documents the construction
of Market-1T; Appendix B details the training and evaluation protocol; Appendix C
presents additional figures, tables, and robustness analyses; and Appendix D
discusses the use of information coefficient as the primary predictive metric.

\section*{Appendix Contents}
\noindent
\hyperref[app_dataset_creation]{\textbf{Appendix A: Dataset Construction Details}}\dotfill \pageref{app_dataset_creation}\\
\hspace*{1.5em}\hyperref[app:why-1hz]{A.1 Why store the market at 1 Hz}\dotfill \pageref{app:why-1hz}\\
\hspace*{1.5em}\hyperref[app:data-source]{A.2 Data source and scope}\dotfill \pageref{app:data-source}\\
\hspace*{1.5em}\hyperref[app:synchronization]{A.3 Synchronization and one-second aggregation}\dotfill \pageref{app:synchronization}\\
\hspace*{1.5em}\hyperref[app:coarser-aggregation]{A.4 Aggregation to coarser resolutions}\dotfill \pageref{app:coarser-aggregation}\\
\hspace*{1.5em}\hyperref[app:sparse-storage]{A.5 Sparse storage and dense reconstruction}\dotfill \pageref{app:sparse-storage}\\
\hspace*{1.5em}\hyperref[app:benchmark-universe]{A.6 Benchmark universe construction}\dotfill \pageref{app:benchmark-universe}\\
\hspace*{1.5em}\hyperref[app:training-normalization]{A.7 Training-time normalization}\dotfill \pageref{app:training-normalization}\\[0.35em]
\hyperref[app:eval]{\textbf{Appendix B: Evaluation}}\dotfill \pageref{app:eval}\\
\hyperref[app:add_res]{\textbf{Appendix C: Additional Results}}\dotfill \pageref{app:add_res}\\
\hyperref[app:ic-vs-sharpe]{\textbf{Appendix D: Choice of Metric: IC vs.\ Sharpe Ratio}}\dotfill \pageref{app:ic-vs-sharpe}

\clearpage
\section{Dataset Construction Details}\label{app_dataset_creation}

This appendix describes the construction of \texttt{Market-1T}, including the raw market feeds, synchronization of asynchronous quotes and trades, trade filtering, sparse storage, and construction of the benchmark universe.

\subsection{Why store the market at 1 Hz}\label{app:why-1hz}

Financial market prediction spans a wide range of timescales. At very low frequencies, variation is increasingly associated with fundamentals, macroeconomic conditions, and longer-run economic developments; at extremely high frequencies, exchange mechanics, latency, queue position, and other microstructure effects become increasingly important. Our primary interest lies between these regimes, where information arrival, market reactions, liquidity, and short-horizon trading dynamics can all matter.

\texttt{Market-1T} therefore stores the market at a base resolution of one second. This resolution is fine enough to preserve short-horizon order-book and trading dynamics while remaining straightforward to aggregate to longer intervals. In particular, all released features admit economically standard aggregation rules, allowing the same data to be represented at second-, minute-, or longer-frequency resolutions without interpolation.

\subsection{Data source and scope}\label{app:data-source}

The raw data are sourced from the \href{https://massive.com}{\texttt{Massive}} flat-file repository and cover U.S.-traded equities from 2008 through 2025. The market data originate from the Securities Information Processor (SIP), specifically the Consolidated Tape Association (CTA) and Unlisted Trading Privileges (UTP) feeds. We ingest two primary event streams:
\begin{itemize}
    \item \textbf{Quotes:} National Best Bid and Offer (NBBO) updates.
    \item \textbf{Trades:} consolidated trade execution reports.
\end{itemize}

We additionally ingest security metadata used for universe construction and filtering, including descriptors such as security type, primary exchange, and share class. The resulting dataset covers the changing universe of equities listed on U.S. exchanges, including securities issued by non-U.S. firms.

\subsection{Synchronization and one-second aggregation}\label{app:synchronization}

The underlying quote and trade feeds are event-driven and therefore asynchronous. We convert them into synchronous one-second intervals. A timestamp $t$ denotes the end of its corresponding interval; for example, the observation stamped 09:30:01 summarizes information from $(09{:}30{:}00, 09{:}30{:}01]$ together with the market state known at 09:30:01.

\subsubsection{Quote state}

Quote variables are state variables rather than flows. For each timestamp $t$, we record the most recent valid NBBO observation available at or before $t$, corresponding to a last-observation-carried-forward (LOCF) representation. Thus,
\[
(P_{bid,t},S_{bid,t},P_{ask,t},S_{ask,t})
\]
represents the quoted market state observable to an agent at the end of the interval rather than an average over quote updates within the interval.

\subsubsection{Trade aggregation}

Trade variables summarize eligible transactions occurring during each one-second interval. We construct:
\begin{itemize}
    \item $P_{open}$: price of the first eligible trade,
    \item $P_{high}$: highest eligible trade price,
    \item $P_{low}$: lowest eligible trade price,
    \item $P_{close}$: price of the last eligible trade,
    \item $P_{vwap}$: volume-weighted average eligible trade price,
    \item $V_{vol}$: total eligible traded volume,
    \item $N_{count}$: number of eligible trades.
\end{itemize}

Together with the four quote-state variables, these produce the 11 released features:
\begin{itemize}
    \item \textbf{Order Book State (4):}
    $P_{bid},S_{bid},P_{ask},S_{ask}$.
    \item \textbf{Price Summary (5):}
    $P_{open},P_{high},P_{low},P_{close},P_{vwap}$.
    \item \textbf{Activity (2):}
    $V_{vol},N_{count}$.
\end{itemize}

\subsubsection{Trade-condition filtering}

Consolidated trade feeds contain condition codes that determine whether a reported transaction should contribute to price and volume statistics. We apply CTA/UTP trade-condition rules before constructing the one-second aggregates. In particular, transactions marked as non-price-forming are excluded from the relevant OHLC statistics, and late or otherwise specially reported transactions are treated according to their eligibility for price and volume calculations. This avoids mechanically incorporating reports that do not represent contemporaneous executable trading activity.

% If you want the exact implementation details exposed in the paper, keep the
% two rules below. Otherwise I would move the exact condition-code mapping to
% released documentation/code because exchange-plan definitions can be subtle.
%
% \begin{itemize}
%     \item \textbf{Price eligibility:} ...
%     \item \textbf{Volume eligibility:} ...
% \end{itemize}

VWAP is computed using the set of trades eligible for the corresponding interval:
\[
P_{vwap}
=
\frac{\sum_{i \in \mathcal{T}} p_i q_i}
     {\sum_{i \in \mathcal{T}} q_i},
\]
where $\mathcal{T}$ denotes eligible trades in the interval, $p_i$ is transaction price, and $q_i$ is transaction size. If no eligible trade occurs during an interval, the corresponding trade-flow variables contain no new transaction information; handling during dense reconstruction is described below.

\subsection{Aggregation to coarser resolutions}\label{app:coarser-aggregation}

Because each feature has a well-defined financial interpretation, one-second observations can be aggregated exactly to coarser intervals. Given a collection of consecutive base intervals, we:
\begin{enumerate}
    \item take the final bid and ask prices and sizes as the ending order-book state;
    \item take the first, maximum, minimum, and final eligible transaction prices as open, high, low, and close;
    \item compute VWAP across all eligible trades in the combined interval; and
    \item sum traded volume and trade count.
\end{enumerate}

These same rules are used when constructing the variable-resolution views in our random scaling and cropping augmentation. Unlike generic interpolation of a time series, this operation preserves the meaning of each market variable after resampling.

\subsection{Sparse storage and dense reconstruction}\label{app:sparse-storage}

Storing a fully materialized record for every security at every second is inefficient because many intervals contain no new quote or trade event. We therefore provide a sparse representation in which intervals with no new market activity need not be physically stored. The dataloader reconstructs the dense 1 Hz sequence as needed.

State variables are carried forward until superseded by a new observation. In particular, bid and ask states persist until the next valid quote update. Flow variables such as traded volume and trade count are zero in intervals containing no eligible trades. Price-summary fields are reconstructed consistently with the downstream representation required by the model; the released dataloaders provide the canonical dense reconstruction used in our experiments.

We additionally support fully dense storage. Dense storage increases disk requirements but can reduce CPU-side reconstruction overhead when data-loading throughput would otherwise bottleneck accelerator utilization.

\subsection{Benchmark universe construction}\label{app:benchmark-universe}

The raw dataset contains the broad universe of U.S.-traded equities. For the representation-learning benchmark, we define a standard high-liquidity common-stock universe intended to reduce the influence of extremely illiquid and non-standard instruments.

The universe is recomputed independently for every month. Eligibility in month $m$ depends only on security metadata and trading statistics measured during month $m-1$, so neither future survival nor future liquidity can affect inclusion. A security must satisfy:
\begin{enumerate}
    \item \textbf{Asset class:} common stock; ETFs, REITs, test securities, and other non-standard instruments are excluded.
    \item \textbf{Price:} prior-month average VWAP $>\$5$.
    \item \textbf{Liquidity:} prior-month average daily dollar volume $\geq \$20\mathrm{M}$.
    \item \textbf{Activity density:} trading activity in more than $20\%$ of regular-trading-hours seconds.
\end{enumerate}

This construction avoids the survivorship bias that would arise from evaluating on a fixed list of firms selected using information from the end of the sample. It also ensures that benchmark membership adapts to listings, delistings, changes in liquidity, and changes in security status through time.

The release includes the metadata necessary to reproduce this benchmark universe and to construct alternative universes using different selection criteria.

\subsection{Training-time normalization}\label{app:training-normalization}

Normalization statistics are computed using training data only. We normalize variables according to their type rather than independently standardizing every channel.

All price variables
\[
P_{bid},P_{ask},P_{open},P_{high},P_{low},P_{close},P_{vwap}
\]
share a common mean and standard deviation. Using common normalization parameters preserves meaningful relative price differences such as the bid--ask spread and the ordering of low, VWAP, and high.

Bid and ask sizes are first transformed as
\[
x \mapsto \log(1+x)
\]
and then jointly standardized. Traded volume and trade count are transformed in the same way and standardized separately. This reduces the influence of the heavy right tails of market-activity variables while preserving the relative scale of variables that have a common economic interpretation.

\clearpage

\section{Evaluation}
\label{app:eval}
As Figure~\ref{fig:multihead_highlighted} illustrates, model performance varies substantially across market regimes. We therefore evaluate every method on the same set of 32 months, sampled throughout the 2008--2025 dataset (Figure~\ref{fig:multihead_highlighted}). For each evaluation month, the model is trained or fitted using only the preceding six months of data. Hyperparameters are selected using a separate set of five months drawn from across the sample.

We standardize the experimental setup across methods wherever possible. All models use a ViT-S backbone. For predictive evaluations, probes are fitted to the final token embedding; for latent-space evaluations, we instead use the mean-pooled representation. The supervised models likewise attach their prediction heads to the final token embedding, whereas self-supervised objectives are applied to the mean-pooled representation. We use fixed sinusoidal positional embeddings for the self-supervised methods, for which they consistently perform well. For supervised training, both sinusoidal embeddings and RoPE perform well, while learned positional embeddings perform substantially worse in our experiments.

We include an `information token' which provides information on the normalization ($\mu$ and $\sigma$ for the four groups of normalization, start time, end time, and aggregation scale in second per token.) This is then mapped into embedding space with a learned 11 $->$ 384 fully connected layer. Our TSFMs do not recieve this token. For our finance baselines, we provide these numbers as features. 

We treat the number of trading days ($\approx 20\times6$) times the number of stocks per day ($\approx 700$) as the number of sampling slots in a six-month training window. The stock universe varies slightly from day to day due to IPOs, delistings, and other changes in universe membership, although in practice the number of eligible stocks is nearly constant within a given six-month period. These slots do not correspond to fixed stock identities: whenever a day is sampled, we draw 16 stocks uniformly at random from the securities available on that day. Thus, an approximately equivalent implementation would be to fix the number of sampling slots per day and compensate with slightly more or fewer training epochs.

Our supervised model is trained with a pairwise learning-to-rank loss. At each optimization step, we sample 256 such slots and, for each, sample 16 stocks from the corresponding day's universe. We make 12 full passes over the training data (scaling results are shown in Figure~\ref{fig:scaling}). For the SSL methods, we approximately match the total number of views processed during training. To evaluate the strength of our supervised models, we compare against time-series foundation models and finance baselines in Table~\ref{tab:finance_baselines}.

For each method, we initialize the configuration from the defaults provided by its official implementation and perform two stages of hyperparameter optimization. In the first stage, we sweep four hyperparameters independently, evaluating four values for each. In the second stage, we identify the two most sensitive hyperparameters and perform a joint grid search of up to $3\times3$ configurations. Hyperparameter selection is performed independently on each of the five tuning months, resulting in up to $(16+9)\times5=125$ optimization runs per method, followed by 32 final evaluation runs. A typical run requires approximately 1--2 H100 GPU-hours.

For LeJEPA, DINO, and BYOL, we use time warping as the default augmentation. We study alternative augmentation choices primarily with LeJEPA, which we found to be the most stable method across configurations. For these augmentation experiments, we perform a single additional sweep over $\lambda$ rather than repeating the full two-stage hyperparameter search.

We consider five strategies for constructing paired self-supervised views:

\begin{itemize}
    \item \textbf{Same Stock Views.} Both views are sampled from the same stock on the same trading day, but use independently sampled crops and transformations. This encourages invariance to the particular temporal subwindow used to represent the same underlying stock-day state.
    \item \textbf{Time Warping.} Both views are drawn from the same stock-day, but are aggregated over different effective temporal scales. Concretely, we randomly compress or dilate the time axis before constructing the model input, so that corresponding views cover different amounts of clock time while preserving the ordering and financial semantics of the underlying observations. The objective therefore encourages the representation to remain stable to moderate changes in temporal scale.
    \item \textbf{Noising.} We add independent noise to the normalized input features of each view, encouraging robustness to small local perturbations in the observed market state.
    \item \textbf{Cross-Stock.} The two views are sampled from different stocks at the same time. This augmentation encourages the representation to retain components of market state that are shared across securities, such as broad market-wide movements, while becoming less sensitive to idiosyncratic stock-level variation. This training objective is most matched to organization task 2, Own-Day Centroid. 
    \item \textbf{Cross-Stock Same Industry.} The two views are sampled from different stocks in the same industry and at the same time, with industries defined using the Fama--French 49-industry classification \citep{FAMA1997153}. Compared with unrestricted cross-stock pairing, this construction places a stronger prior on shared exposures and common economic structure within the positive pair. This training objective is most matched to organization task 1, partner matched view.
\end{itemize}

\clearpage

\section{Additional Results}
\label{app:add_res}

\begin{figure}[h]
    \centering
    \includegraphics[width=1\linewidth]{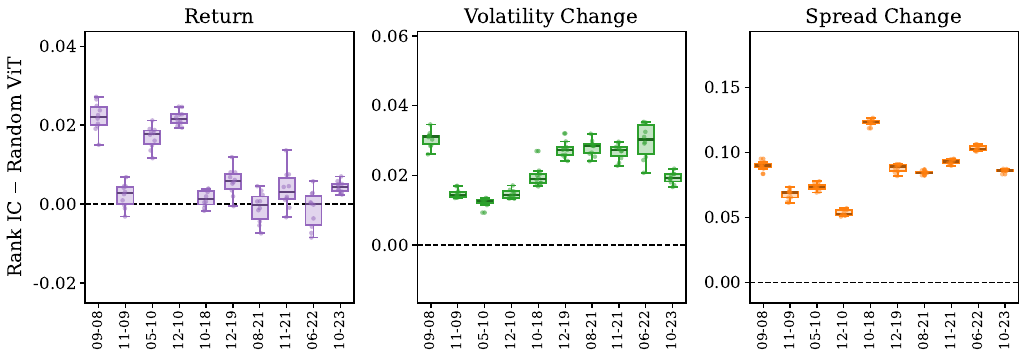}
    \caption{If we subtract off the Random ViT baseline for each month, we can substantially reduce the amount of variation that is from the choice of month, allowing for better comparisons. In practice, we can use this to adjust our standard error calculation or make a more fair comparison between performance on different months.}
    \label{fig:variance_decomp_app}
\end{figure}

\begin{figure}
    \centering
    \includegraphics[width=1\linewidth]{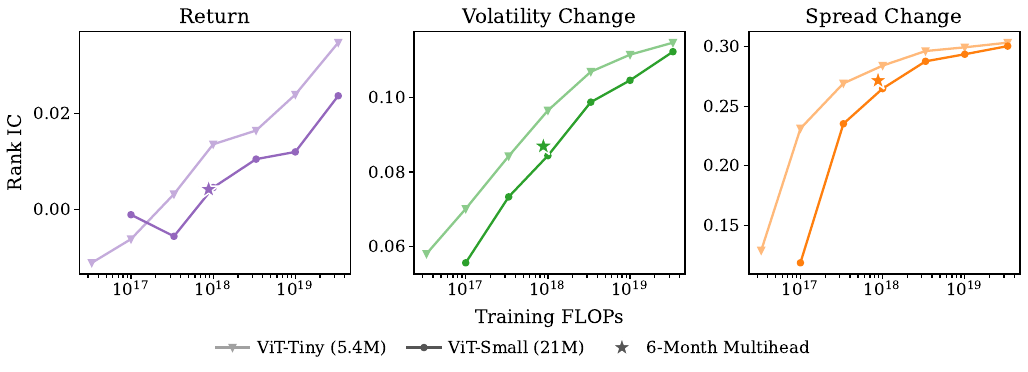}
    \caption{We train a single supervised multihead encoder on 2008-2017 and evaluated it on January 2018, compared to a standard multihead encoder. This run goes takes the top-end from ViT-Tiny from $3.33*10^{17}$ to $3.33*10^{19}$ FLOPs as compared to Figure \ref{fig:scaling}. Since the multihead encoder normalizes the gradient between all three tasks, it uses 3$\times$ the FLOPs per data sample. We see that for return and volatility change the model does not yet saturate.}
    \label{fig:scaling_decade}
\end{figure}

\begin{table}[t]
\caption{Forecasting rank IC at the 15-minute horizon for classical baselines, beside the supervised arms of Table~\ref{tab:probe_fit}, averaged over evaluation months. The baselines are read off the \emph{normalised view tensor the encoder receives}, not the raw order book. Based on the results on the Optimization Set, the frozen TSFMs were nearly uniformly optimal at the last layer for all three tasks. Supervised cells give the ridge probe with that arm's own head IC in parentheses. \textcolor{green!45!black}{\textbf{Green}} marks the best two strongest supervised arms and the two strongest baselines.}
\label{tab:finance_baselines}
\centering
\begin{tabular}{lccc}
\toprule
 & Return & Volatility & Spread \\
\midrule
\addlinespace \multicolumn{4}{l}{\emph{Supervised}} \\
Return & \textcolor{green!45!black}{$\mathbf{0.0272}$}\,($0.0269$) & $0.0695$ & $0.1305$ \\
Vol & $0.0197$ & \textcolor{green!45!black}{$\mathbf{0.0883}$}\,($0.0891$) & $0.1480$ \\
Spread & $0.0178$ & $0.0762$ & \textcolor{green!45!black}{$\mathbf{0.2517}$}\,($0.2519$) \\
Multihead & \textcolor{green!45!black}{$\mathbf{0.0311}$}\,($0.0317$) & \textcolor{green!45!black}{$\mathbf{0.0941}$}\,($0.0952$) & \textcolor{green!45!black}{$\mathbf{0.2455}$}\,($0.2441$) \\
\addlinespace \multicolumn{4}{l}{\emph{Classical forecasting models}} \\
Mean reversion & -- & $0.0274$ & $0.1031$ \\
AR(p) & $0.0136$ & $0.0442$ & $0.1688$ \\
ARMA(p,q) & $0.0141$ & $0.0455$ & $0.1740$ \\
HAR-RV & -- & $0.0639$ & -- \\
GARCH(1,1) & -- & $0.0468$ & -- \\
Ridge ARDL & $0.0180$ & $0.0624$ & \textcolor{green!45!black}{$\mathbf{0.1857}$} \\
\addlinespace \multicolumn{4}{l}{\emph{Learners on the raw view}} \\
Ridge (8-token tail) & $0.0175$ & $0.0667$ & $0.1507$ \\
Ridge (24-token tail) & $0.0196$ & $0.0707$ & $0.1660$ \\
Ridge (64-token tail) & $0.0178$ & $0.0709$ & \textcolor{green!45!black}{$\mathbf{0.1820}$} \\
GBM (24-token tail) & $0.0097$ & $0.0632$ & $0.1034$ \\
\addlinespace \multicolumn{4}{l}{\emph{Frozen time-series foundation models}} \\
Chronos-2 & \textcolor{green!45!black}{$\mathbf{0.0230}$} & \textcolor{green!45!black}{$\mathbf{0.0928}$} & $0.1289$ \\
Kronos & $0.0171$ & $0.0621$ & $0.0560$ \\
TimesFM 3.0 & \textcolor{green!45!black}{$\mathbf{0.0232}$} & \textcolor{green!45!black}{$\mathbf{0.0947}$} & $0.1362$ \\
\addlinespace \multicolumn{4}{l}{\emph{Untrained floor}} \\
Random ViT & $0.0170$ & $0.0694$ & $0.1692$ \\
\bottomrule
\end{tabular}
\end{table}

\clearpage

\begin{figure}
    \centering
    \includegraphics[width=1\linewidth]{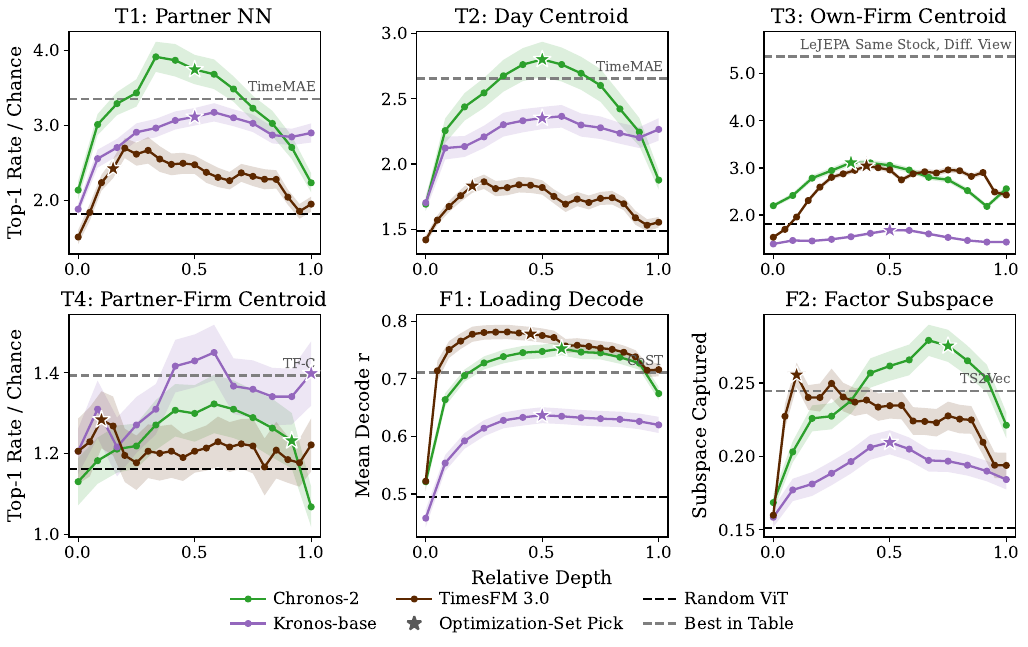}
    
    \vspace{1em}
    
    \includegraphics[width=1\linewidth]{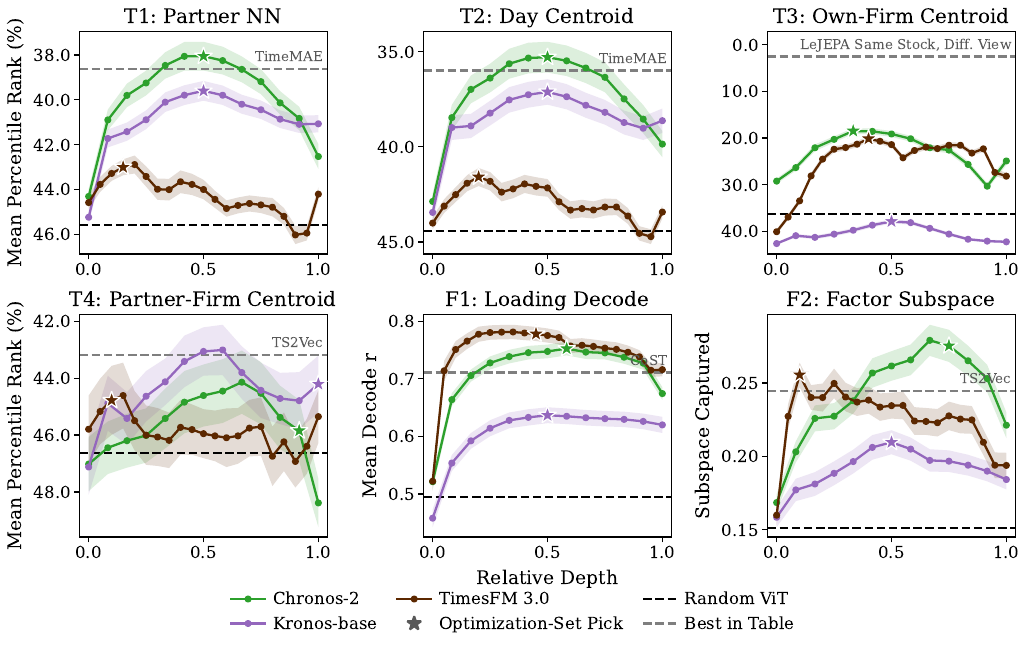}
    \caption{Sweeping all layers for the latent-space evaluations for the TSFMs. Stars show the optimal layer on the five month optimization set. We note that is not an entirely fair comparison, as all other methods are only presented on the last layer, and that other methods (both the default supervised models, and the SSL methods) will likely also have internal layers where performance is stronger than the last layer for each task.}
    \label{fig:tsfm_latent_sweep}
\end{figure}
\clearpage
\begin{table}[t]
\caption{RankMe~\citep{garrido2023rankmeassessingdownstreamperformance} of each frozen embedding on the latent suite's full-day panels of Table~\ref{tab:latent_results}, read at the same mean pooling, mean $\pm$ SE over the eval months, beside the embedding width that bounds it. RankMe counts how many directions the embedding spends its variance on; it is not a score -- the supervised models, the strongest forecasters, have the lowest -- and a width far from 384 puts a row off the rest of the column's scale. The frozen TSFMs are read at their last layer with the nine per-channel states averaged, the prediction evals' readout; Figure~\ref{fig:rankme_sweep} sweeps every layer.}
\label{tab:rankme_latent}
\centering
\begin{tabular}{lcc}
\toprule
 & Width & RankMe \\
\midrule
\addlinespace \multicolumn{3}{l}{\emph{Supervised}} \\
Return & $384$ & $2.8 \pm 0.1$ \\
Vol & $384$ & $3.0 \pm 0.0$ \\
Spread & $384$ & $3.5 \pm 0.2$ \\
Multihead & $384$ & $3.7 \pm 0.1$ \\
\addlinespace \multicolumn{3}{l}{\emph{LeJEPA}~\citep{balestriero2025lejepaprovablescalableselfsupervised}} \\
\scriptsize{Same Stock, Diff. View} & $384$ & $11.1 \pm 0.1$ \\
Time Warping & $384$ & $14.3 \pm 0.2$ \\
Gaussian Noising & $384$ & $12.4 \pm 0.1$ \\
Cross Stock & $384$ & $14.8 \pm 0.1$ \\
C-S, Same Industry & $384$ & $15.3 \pm 0.1$ \\
\addlinespace \multicolumn{3}{l}{\emph{SSL}} \\
DINO~\citep{caron_etal_dino} & $384$ & $11.2 \pm 0.1$ \\
BYOL~\citep{grill2020bootstraplatentnewapproach} & $384$ & $10.9 \pm 0.2$ \\
CPC~\citep{oord2019representationlearningcontrastivepredictive} & $384$ & $13.0 \pm 0.4$ \\
I-JEPA~\citep{assran2023selfsupervisedlearningimagesjointembedding} & $384$ & $5.5 \pm 0.1$ \\
MAE~\citep{he_etal_2022} & $384$ & $10.6 \pm 0.8$ \\
TS2Vec~\citep{yue2022ts2vec} & $384$ & $6.8 \pm 0.1$ \\
CoST~\citep{woo2022costcontrastivelearningdisentangled} & $384$ & $17.3 \pm 0.3$ \\
TF-C~\citep{zhang2022selfsupervisedcontrastivepretrainingtime} & $256$ & $81.4 \pm 1.1$ \\
TimeMAE~\citep{Cheng_2026} & $384$ & $61.7 \pm 0.6$ \\
\addlinespace \multicolumn{3}{l}{\emph{Frozen TSFMs (last layer, channels averaged)}} \\
Chronos-2 & $768$ & $107.1 \pm 0.6$ \\
Kronos & $832$ & $77.3 \pm 0.5$ \\
TimesFM 3.0 & $1280$ & $111.9 \pm 0.6$ \\
\addlinespace \multicolumn{3}{l}{\emph{Untrained floor}} \\
Random ViT & $384$ & $43.2 \pm 0.1$ \\
\bottomrule
\end{tabular}
\end{table}

\begin{figure}
    \centering
    \includegraphics[width=0.5\linewidth]{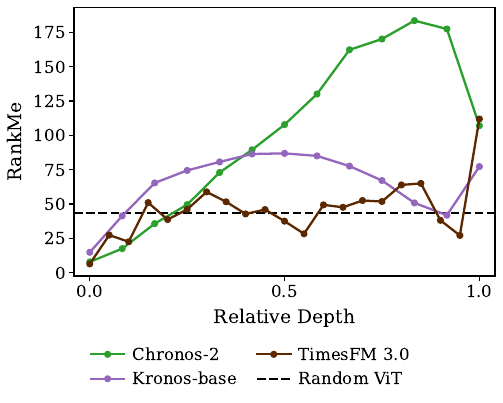}
    \caption{Sweep of RankMe across TSFM layers. $\pm$ 1 SE bands are narrower than the line width.}
    \label{fig:rankme_sweep}
\end{figure}
%\FloatBarrier
\clearpage
\section{Choice of Metric: IC vs.\ Sharpe Ratio}
\label{app:ic-vs-sharpe}

\begin{figure}[h]
    \centering
    \includegraphics[width=1\linewidth]{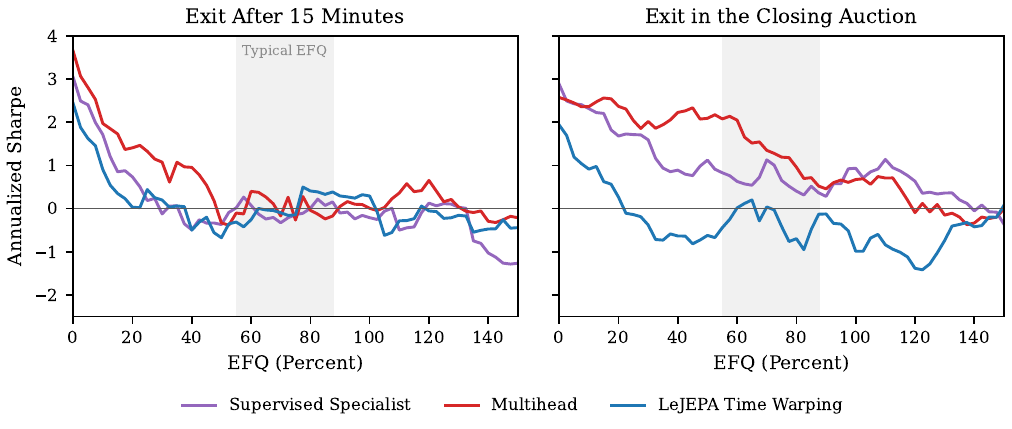}
    \caption{Annualized Sharpe of a dollar-neutral cross-sectional book as a
function of EFQ, the effective-to-quoted half-spread ratio, where $0\%$ is a
midpoint fill and $100\%$ pays the entire quoted spread. Once per day at
13:45~ET we take every name whose forecast edge exceeds the round trip it
would actually be charged at that EFQ, equal-weight the long and short legs,
and either unwind after fifteen minutes (left) or hold to the 16:00~ET
closing auction (right); the shaded band is the EFQ range \cite{levy2022price}
measures for orders routed to the market rather than sold to a wholesaler,
$55\%$ for direct market access up to $88\%$ on NYSE. We sweep past $100\%$
because an order larger than the quoted depth walks the book, and because a
signal others are trading moves the quote away before the order completes.}
    \label{fig:efq}
\end{figure}

For evaluating predictive information in the representation, we prefer the information coefficient (IC) to the Sharpe ratio as our primary benchmark. Firstly, even weak-form market efficiency \citep{7104d0bd-69aa-330c-badb-94aa0c8f5bac} suggests 
that using only price and volume will not be profitable to trade (strong-form market efficiency
suggests that even with all available information it is not possible.) 
This does not require that a learned representation carry no signal but that the signal be no larger than the cost of acting on it, which for a high-frequency cross-sectional forecast is the bid-ask spread. We evaluate as a function of EFQ, the effective-to-quote half spread ratio, with simple strategies (e.g. lower bounds on Sharpe) varied to try to reduce trading costs (large amount of liquidity at closing auction). See Figure \ref{fig:efq}

% Generated by plots/ic_vs_sharpe/ic_sharpe_deliverables.py -- do not edit.
% Edit OPTION_LABELS / AXIS_HEADERS / CAPTION in that script instead;
% this file is overwritten wholesale on every run.
\begin{table}[h]
  \centering
  \footnotesize
  \setlength{\tabcolsep}{6pt}
  \begin{tabular}{llllll}
  \toprule
  \textbf{Universe} & \textbf{Selection} & \textbf{Weighting} & \textbf{Risk model} & \textbf{EFQ} & \textbf{Rebalance} \\
  \midrule
  All & All & Equal & Diagonal & $0\%$ & Every decision \\
  Tightest 50\% & Quintile & Inverse vol. & Shrunk sample & $25\%$ & Daily \\
  Tightest 25\% & Decile & Min.\ variance& Ledoit--Wolf~\citep{LEDOIT2003603} & $50\%$ &  \\
   & Long-only & ERC~\citep{maillard2010properties} & Embedding & $70\%$ &  \\
   &  & Rank &  & $88\%$ &  \\
   &  & Mean--variance~\citep{e5a1bb8f-41b7-35c6-95cd-8b366d3e99bc} &  & $100\%$ &  \\
  \bottomrule
  \end{tabular}
  \caption{Six of the many choices that lie between a forward return
    prediction and a Sharpe ratio, and the options we sweep at each. Their
    product, less combinations that are degenerate by construction, gives
    the $2{,}448$ configurations we use to demonstrate how wide a spread
    these choices induce. Execution quality is EFQ, the effective half-spread
    actually paid over the half-spread quoted when the order was sent
    \citep{levy2022price}, at $0\%$ a midpoint fill and at $100\%$ the whole
    quoted spread.}
  \label{tab:ic-sharpe-choices}
\end{table}
% 2448 distinct configurations after removing degenerate pairs

% Generated by plots/ic_vs_sharpe/ic_sharpe_deliverables.py -- do not edit.
% Edit HARNESS_CAPTION / AXIS_LABELS in that script instead;
% this file is overwritten wholesale on every run.
\begin{table}[t]
  \centering
  \footnotesize
  \setlength{\tabcolsep}{4pt}
  \begin{tabular}{l rr @{\hspace{14pt}} rrrrr}
  \toprule
  & \multicolumn{2}{c}{\textbf{IC vs.\ Sharpe}} & \multicolumn{5}{c}{\textbf{Sharpe range across each choice}} \\
  \cmidrule(lr){2-3} \cmidrule(l){4-8}
  \textbf{EFQ} & $r(\mathrm{IC},\bar{S})$ & $\bar{r}(\mathrm{IC},S)$ & \textbf{Rebalance} & \textbf{Selection} & \textbf{Universe screen} & \textbf{Weighting} & \textbf{Risk model} \\
  \midrule
  $0\%$ & +0.92 & +0.52 & 2.3 & 1.6 & 1.6 & 2.4 & 0.4 \\
  $25\%$ & +0.90 & +0.49 & 4.9 & 2.3 & 2.2 & 1.4 & 1.4 \\
  $50\%$ & +0.77 & +0.41 & 11.9 & 6.1 & 5.7 & 1.3 & 2.2 \\
  $70\%$ & +0.64 & +0.36 & 17.2 & 9.1 & 8.4 & 2.9 & 2.7 \\
  $100\%$ & +0.46 & +0.29 & 24.7 & 13.3 & 11.9 & 5.5 & 3.4 \\
  \bottomrule
  \end{tabular}
  \caption{A Sharpe ratio is a property of the harness as much as of
    the forecast, and both halves of this table are read down the same axis of
    execution quality. EFQ is the effective half-spread actually paid over the
    half-spread quoted when the order was sent \citep{levy2022price}, at $0\%$ a
    midpoint fill and at $100\%$ the whole quoted spread.
    \textbf{Left:} $r(\mathrm{IC},\bar{S})$ correlates each method's IC with
    its Sharpe averaged over every harness at that execution quality;
    $\bar{r}(\mathrm{IC},S)$ correlates them inside one harness and then
    averages that correlation over harnesses. $r(\mathrm{IC},\bar{S}) > \bar{r}(\mathrm{IC},S)$ \textbf{Right:} how far each remaining choice moves the
    annualized Sharpe, over 18 methods and the 408 harnesses at
    each execution quality. The choice of
    backtest moves the score much more than the choice of model.}
  \label{tab:ic-sharpe-harness}
\end{table}

Secondly, a Sharpe ratio is jointly a property of the forecast and of the backtest used to turn that forecast into trades. Portfolio construction, security selection, risk modeling, rebalance frequency, and assumptions about execution all change the realized strategy even when the underlying forecast is held fixed. Our harness therefore varies these choices explicitly rather than treating any one implementation as canonical.

Across methods, IC is much more closely related to performance after these implementation choices are averaged away than it is within any particular backtest. At the same time, changing the backtest specification can move the resulting Sharpe substantially more than changing the encoder. Thus, Sharpe is useful for demonstrating whether a forecast can survive realistic implementation choices, but is a less stable benchmark for comparing representations themselves. We therefore use IC as the primary predictive metric and treat the portfolio experiments as a downstream robustness check.

\clearpage

\clearpage
\section*{NeurIPS Paper Checklist}

\begin{enumerate}

\item {\bf Claims}
    \item[] Question: Do the main claims made in the abstract and introduction accurately reflect the paper's contributions and scope?
    \item[] Answer: \answerYes{}
    \item[] Justification: Yes, the abstract and introduction discuss a dataset, library, and study of representation learning in finance as a foundation for world modeling.

\item {\bf Limitations}
    \item[] Question: Does the paper discuss the limitations of the work performed by the authors?
    \item[] Answer: \answerYes{}
    \item[] Justification: Yes, the paper discusses that the evaluations are limited in scope. 

\item {\bf Theory assumptions and proofs}
    \item[] Question: For each theoretical result, does the paper provide the full set of assumptions and a complete (and correct) proof?
    \item[] Answer: \answerNA{} % Replace by \answerYes{}, \answerNo{}, or \answerNA{}.
    \item[] Justification: No theoretical proofs. 

    \item {\bf Experimental result reproducibility}
    \item[] Question: Does the paper fully disclose all the information needed to reproduce the main experimental results of the paper to the extent that it affects the main claims and/or conclusions of the paper (regardless of whether the code and data are provided or not)?
    \item[] Answer: \answerYes{}
    \item[] Justification: Yes, all the information and code is provided.

\item {\bf Open access to data and code}
    \item[] Question: Does the paper provide open access to the data and code, with sufficient instructions to faithfully reproduce the main experimental results, as described in supplemental material?
    \item[] Answer: \answerYes{}
    \item[] Justification: The dataset and pre-trained encoders are available at
\href{https://huggingface.co/fin-ai-lab}{\textcolor{blue}{https://huggingface.co/fin-ai-lab}}.
The code is available at
\href{https://github.com/fin-ai-lab/tfwm}{\textcolor{blue}{https://github.com/fin-ai-lab/tfwm}}.

\item {\bf Experimental setting/details}
    \item[] Question: Does the paper specify all the training and test details (e.g., data splits, hyperparameters, how they were chosen, type of optimizer) necessary to understand the results?
    \item[] Answer: \answerYes{}
    \item[] Justification: All information is included. 

\item {\bf Experiment statistical significance}
    \item[] Question: Does the paper report error bars suitably and correctly defined or other appropriate information about the statistical significance of the experiments?
    \item[] Answer: \answerYes{}
    \item[] Justification: Yes, all plots have 2 SE bar. 

\item {\bf Experiments compute resources}
    \item[] Question: For each experiment, does the paper provide sufficient information on the computer resources (type of compute workers, memory, time of execution) needed to reproduce the experiments?
    \item[] Answer: \answerNo{}
    \item[] Justification: The paper does not estimate the cost of running all experiments. 
    
\item {\bf Code of ethics}
    \item[] Question: Does the research conducted in the paper conform, in every respect, with the NeurIPS Code of Ethics \url{https://neurips.cc/public/EthicsGuidelines}?
    \item[] Answer: \answerYes{}
    \item[] Justification: The authors have reviewed the code of ethics. 

\item {\bf Broader impacts}
    \item[] Question: Does the paper discuss both potential positive societal impacts and negative societal impacts of the work performed?
    \item[] Answer: \answerNA{}
    \item[] Justification: There are no societal impacts of this work. 
    
\item {\bf Safeguards}
    \item[] Question: Does the paper describe safeguards that have been put in place for responsible release of data or models that have a high risk for misuse (e.g., pre-trained language models, image generators, or scraped datasets)?
    \item[] Answer: \answerNA{}
    \item[] Justification: Poses no such risks. 

\item {\bf Licenses for existing assets}
    \item[] Question: Are the creators or original owners of assets (e.g., code, data, models), used in the paper, properly credited and are the license and terms of use explicitly mentioned and properly respected?
    \item[] Answer: \answerYes{}
    \item[] Justification: Yes, the dataset is fully licensed for release. 

\item {\bf New assets}
    \item[] Question: Are new assets introduced in the paper well documented and is the documentation provided alongside the assets?
    \item[] Answer: \answerYes{}
    \item[] Justification: Yes, the dataset is well documented. 

\item {\bf Crowdsourcing and research with human subjects}
    \item[] Question: For crowdsourcing experiments and research with human subjects, does the paper include the full text of instructions given to participants and screenshots, if applicable, as well as details about compensation (if any)? 
    \item[] Answer: \answerNA{}.
    \item[] Justification: No crowdsourcing / research with human subjects. 

\item {\bf Institutional review board (IRB) approvals or equivalent for research with human subjects}
    \item[] Question: Does the paper describe potential risks incurred by study participants, whether such risks were disclosed to the subjects, and whether Institutional Review Board (IRB) approvals (or an equivalent approval/review based on the requirements of your country or institution) were obtained?
    \item[] Answer: \answerNA{}.
    \item[] Justification: No crowdsourcing / research with human subjects. 

\item {\bf Declaration of LLM usage}
    \item[] Question: Does the paper describe the usage of LLMs if it is an important, original, or non-standard component of the core methods in this research? Note that if the LLM is used only for writing, editing, or formatting purposes and does \emph{not} impact the core methodology, scientific rigor, or originality of the research, declaration is not required.
    %this research? 
    \item[] Answer: \answerNA{}
    \item[] Justification: The core method development in this research does not involve LLMs as any important, original, or non-standard components. 

\end{enumerate}

\end{document}